\pdfoutput=1
\documentclass[11pt]{article}

\usepackage[]{acl}

\usepackage{times}
\usepackage{latexsym}

\usepackage{xspace,mfirstuc,tabulary}
\usepackage{booktabs}
\usepackage{amssymb}
\usepackage{amsmath}
\usepackage{multirow,booktabs, hhline}
\usepackage[ruled,noend]{algorithm2e}
\usepackage{amsmath, bm}
\newcommand\ourmodel{IPT\xspace}
\usepackage{graphicx}
\usepackage{color}
\usepackage{subfigure}
\usepackage{enumitem}

\usepackage[T1]{fontenc}
\usepackage[utf8]{inputenc}

\usepackage{cleveref}
\crefname{section}{§}{§§}
\Crefname{section}{§}{§§}
\usepackage{microtype}

\title{Exploring Universal Intrinsic Task Subspace via Prompt Tuning}
\author{
 Yujia~Qin$^1\thanks{* The first two authors contributed equally.}$\hspace{0.5em}, Xiaozhi~Wang$^{1*}$, Yusheng~Su$^1$, Yankai~Lin$^2$, Ning~Ding$^1$, Jing~Yi$^1$, \\
 \textbf{Weize~Chen$^1$, Zhiyuan~Liu$^1$, Juanzi~Li$^1$, Lei~Hou$^1$, Peng~Li$^1$, Maosong~Sun$^1$, Jie~Zhou$^2$} \\
 $^1$Department of Computer Science and Technology, Tsinghua University, Beijing, China \\
 $^2$Pattern Recognition Center, WeChat AI, Tencent Inc. \\
 
\texttt{yujiaqin16@gmail.com}, \texttt{wangxz20@mails.tsinghua.edu.cn}\\
}

\begin{document}
\maketitle

\begin{abstract}
Why can pre-trained language models (PLMs) learn universal representations and effectively adapt to broad NLP tasks differing a lot superficially? In this work, we empirically find evidence indicating that the adaptations of PLMs to various few-shot tasks can be reparameterized as optimizing only a few free parameters in a unified low-dimensional \textit{intrinsic task subspace}, which may help us understand why PLMs could easily adapt to various NLP tasks with small-scale data. To find such a subspace and examine its universality, we propose an analysis pipeline called \textit{intrinsic prompt tuning} (\ourmodel). Specifically, we resort to the recent success of prompt tuning and decompose the soft prompts of multiple NLP tasks into the same low-dimensional nonlinear subspace, then we learn to adapt the PLM to unseen data or tasks by only tuning parameters in this subspace. In the experiments, we study diverse few-shot NLP tasks and surprisingly find that in a $250$-dimensional subspace found with $100$ tasks, by only tuning $250$ free parameters, we can recover $97\%$ and $83\%$ of the full prompt tuning performance for $100$ seen tasks (using different training data) and $20$ unseen tasks, respectively, showing great generalization ability of the found intrinsic task subspace. Besides being an analysis tool, \ourmodel could further help us improve the prompt tuning stability. The codes are publicly available at \url{https://github.com/thunlp/Intrinsic-Prompt-Tuning}.
% we can surprisingly achieve non-trivial performance for both seen and unseen tasks through learning only in the extremely low-dimensional subspace.
% $89\%$ of the full-parameter fine-tuning performance could be achieved in as low as $5$-dimensional a subspace. In addition, we demonstrate the generalization ability of the found intrinsic subspace on unseen data and tasks.
% by only training $5$ parameters in the $5$-dimensional subspace found by decomposing soft prompts of $100$ tasks, we can achieve $87\%$ and $65\%$ of the full soft prompting performance for $100$ seen tasks and $20$ unseen tasks, respectively.
% We name this method to find low-dimensional intrinsic subspaces by decomposing multiple soft prompts and then train new tasks in the intrinsic subspace as \textit{intrinsic prompt tuning} (IPT).
\end{abstract}
\section{Introduction}
\begin{figure}[!t]
\centering
\includegraphics[width=0.45\textwidth]{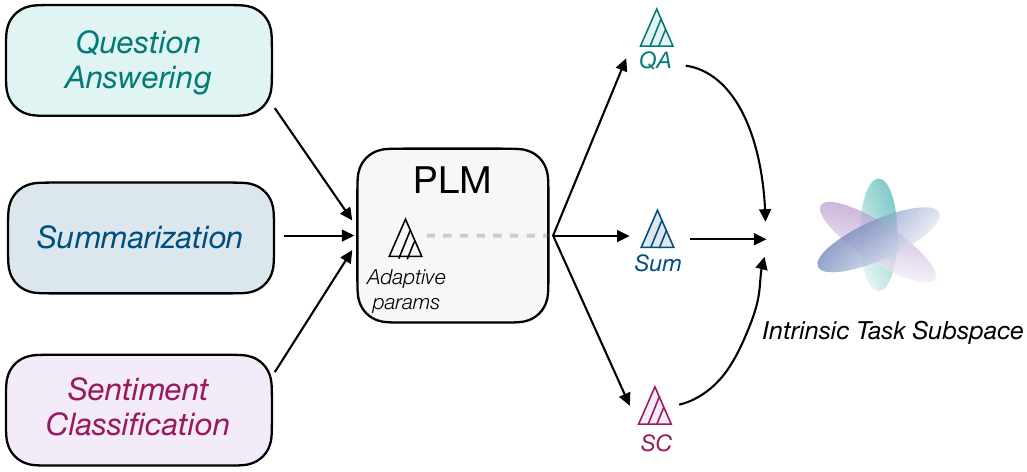}
\caption{An illustration of a common low-dimensional intrinsic task subspace for diverse tasks. PLMs tune adaptive parameters to adapt to each task.}
\label{fig:ipt}
\end{figure}

Pre-trained language models (PLMs) have shown dominant performances on various natural language processing (NLP) tasks~\citep{Han2021PreTrainedMP,min2021recent}. After pre-training huge parameters on massive data, a PLM can effectively adapt to diverse downstream NLP tasks with small-scale data through full-parameter fine-tuning or parameter-efficient tuning methods~\citep{lester2021power,houlsby2019parameter}. Nevertheless, the mechanisms behind such adaptations remain unclear. Why can PLMs learn universal representations through task-irrelevant pre-training objectives and easily adapt to diverse NLP tasks differing a lot? 
%A deep understanding of this question may impact the designing of pre-training and downstream adaption algorithms profoundly.
Towards answering this question, in this paper, we hypothesize that the adaptations of PLMs to various downstream tasks can be reparameterized as optimizing only a few free parameters in a unified low-dimensional parameter subspace, which we call \textit{intrinsic task subspace} (Figure~\ref{fig:ipt}).
% this paper investigates the existence of \textit{a low-dimensional subspace for a broad scope of NLP tasks conditioned on one PLM}. 
% That is, as illustrated in Figure~\ref{fig:ipt}, PLMs learn a re-parameterization $\Delta$ for the tuned parameters during adaptation towards a series of NLP tasks, we explore whether these $\Delta$s could be compressed into low-dimensional representations in a common subspace, which is dubbed as \textit{intrinsic subspace}.  

%to solve a series of NLP problems.
%where a a series NLP tasks could be solved by only manipulating a few parameters.

Specifically, during adaptation to a certain downstream task, PLMs optimize the tunable \textit{adaptive parameters}. This is typically a high-dimensional optimization problem. For instance, in conventional fine-tuning, the adaptive parameters are all the PLM parameters, which may exceed hundreds of millions. However, \citet{aghajanyan-etal-2021-intrinsic} show that the adaptation to a single task of a PLM can be reparameterized into only optimizing hundreds of free parameters in a low-dimensional subspace and then randomly projecting the tuned parameters back into the full parameter space. This motivates our hypothesis that adaptations to multiple tasks can be reparameterized into optimizations in \textbf{a unified} low-dimensional intrinsic task subspace. If this hypothesis holds, then (1) the existence of a common task reparameterization subspace explains the universality of PLMs and (2) the low dimensionality explains why the adaptations can be done with relatively small-scale data. From this perspective, the PLMs serve as general \textit{compression frameworks}, which compress the learning complexity of various tasks from very high dimensionalities to low dimensionalities.

To find evidence for the hypothesis, we need to develop methods for finding the common intrinsic task subspaces of PLMs. Naturally, the subspace should contain adaptation solutions (i.e., tuned adaptive parameters) for various tasks, hence we can approximate the subspace by training a low-dimensional decomposition of the adaptive parameters using multiple tasks and then examine whether we can learn unseen tasks in the found subspace. However, training a decomposition for all the PLM parameters (the case of fine-tuning) is computationally unaffordable since the required parameters of the decomposition would be hundreds of times of PLMs. Fortunately, prompt tuning (PT) provides a parameter-efficient alternative, whose number of adaptive parameters (\textit{soft prompts}), are only tens of thousands. PT can also achieve close performance to fine-tuning on both understanding~\cite{lester2021power} and generation~\cite{li-liang-2021-prefix} tasks. % Moreover, PT does not have structural biases since the tuned soft prompts are confined to input embeddings, hence decomposing them is intuitively easier than other parameter-efficient tuning methods like adapter~\cite{houlsby2019parameter}.

% \dn{the few-shot setting comes out of nowhere}

% The few-shot setting ensures the data scales of various tasks are balanced, so that the subspace found by MSF will not be easily biased towards data-rich tasks.

In experiments, we explore the common intrinsic subspace through PT under the few-shot learning setting, which ensures the data scales of various tasks are balanced. We name the analysis pipeline used in this paper as \textbf{I}ntrinsic \textbf{P}rompt \textbf{T}uning (\textbf{\ourmodel}), which consists of two phases: multi-task subspace finding (MSF) and intrinsic subspace tuning (IST). During MSF, we first obtain trained soft prompts for multiple tasks and then learn an auto-encoder by first projecting them into the desired low-dimensional subspace and then reconstructing them with a back-projection. % The optimized auto-encoder defines the desired intrinsic subspace.
% with a projection function to project the soft prompts into the subspace and a back-projection function to project vectors in the subspace back. Both projection functions are parameterized with nonlinear neural networks so that can be trained with gradient descent algorithms. Then we take optimized prompts as inputs and train the two projection functions with the reconstruction loss and specific losses for training tasks.
% Finally, the subspace defined by the two optimized projections is the desired approximate intrinsic subspace since it compresses multiple training tasks.
During IST, to adapt the PLM to unseen data and tasks, we only train the few free parameters in the low-dimensional subspace found by MSF through a fixed back-projection.

Surprisingly, we find that the intrinsic task subspace may not only exist but also is extremely low-dimensional. We study diverse few-shot NLP tasks and find that in a $250$-dimensional subspace found by $100$ tasks with MSF, we can recover $97\%$ and $83\%$ of the full PT performance with IST for $100$ seen tasks (using different training data) and $20$ unseen tasks, respectively. Furthermore, we analyze the effect of training task types, the number of training tasks, and training data scales for \ourmodel. We also show that \ourmodel and the intrinsic task subspace could help us analyze task differences and improve training stability. We encourage future work to explore how to better find the intrinsic task subspace and develop techniques taking inspiration from universal reparameterizations of PLM adaptations.
\section{Related Work}
\paragraph{PLM, Fine-tuning and Prompt tuning.}
Since the success of BERT~\cite{devlin2018bert}, pre-trained language models bring a new paradigm to NLP, that is to pre-train a massive model as the universal backbone and then adapt the PLMs to specific downstream tasks. The mainstream way of downstream adaptation is fine-tuning, which adds task-specific classification heads and tunes all the PLM parameters with supervised data.

Recently, researchers found that promising results can be achieved by casting downstream tasks into the form of pre-training tasks and adding some \textit{prompt} tokens into the input, including human-designed explainable prompts~\citep{NEURIPS2020_1457c0d6,schick-schutze-2021-exploiting,schick-schutze-2021-just} and automatically searched prompts~\citep{jiang-etal-2020-know,shin-etal-2020-autoprompt,gao-etal-2021-making}. Following this line of study, the prompts are extended from real tokens to trainable embeddings, i.e., soft prompts~\citep{hambardzumyan2021warp,zhong-etal-2021-factual,qin-eisner-2021-learning}. Furthermore, some works~\citep{lester2021power,li-liang-2021-prefix} demonstrate that only tuning soft prompts and keeping PLMs frozen can achieve excellent performance in various tasks, especially for large-scale PLMs. % Especially, \citet{lester2021power} show that with the growth of PLM's size, the gap between prompt tuning and fine-tuning becomes narrower and finally disappears.
In this work, we try to understand these phenomena, i.e., why can PLMs learn universal abilities to adapt to various tasks with few data points and tunable parameters.

\begin{figure*}[!t]
\centering
\includegraphics[width=0.95\textwidth]{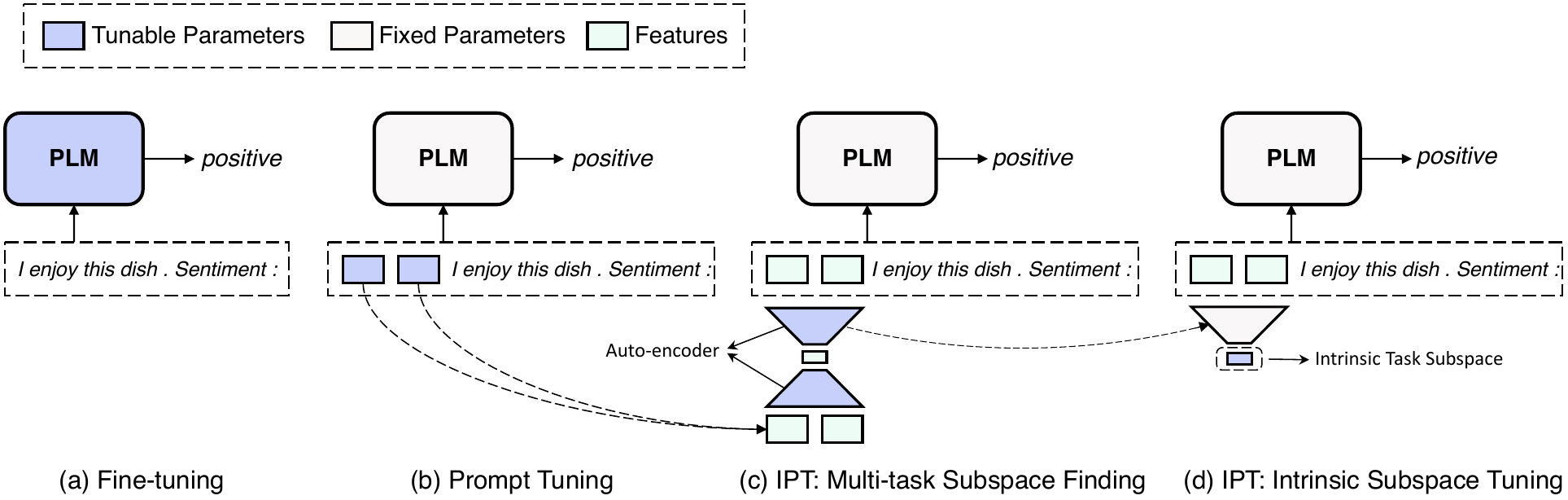}
\caption{Illustrations of (a) fine-tuning, (b) prompt tuning and (c,d) two components of \ourmodel. We discriminate tunable parameters, fixed parameters and intermediate features with different colors.
}
\label{fig:method}
\end{figure*}

\paragraph{Intrinsic Dimensionality.} \textit{Intrinsic dimension} (ID) is the minimal number of variables needed to represent some data or approximate a function. \citet{li2018measuring} propose to measure the IDs of objective functions optimized by neural networks through randomly projecting all the trainable parameters into linear subspaces and finding the minimal dimensions that satisfactory solutions appear. Following this, \citet{aghajanyan-etal-2021-intrinsic} show that the IDs of PLM adaptations (via fine-tuning) to a single task can be smaller than thousands and pre-training implicitly lowers the IDs of downstream tasks, which motivates this work. % The random linear projection used in their methods does not involve any additional training for finding low-dimensional subspaces, so that successfully finding solutions in the subspaces provides ample evidence for the existence of effective low-dimensional reparameterizations.
Considering the existence of individual subspace for each task has been proved, here we aim to study whether the subspace is universal. However, the random linear projections of previous methods inevitably introduce redundant task-irrelevant information and make the investigated subspace not compact for reparameterizing task adaptations. Therefore, we resort to stronger subspace-finding methods and use supervision from diverse tasks to train a nonlinear low-dimensional decomposition for the adaptive parameters.
% The MSF in our work can be partly seen as finding a lower bound for IDs in prompt tuning for PLMs, while we mainly focus on whether the subspaces compressing solutions are shared by different tasks.

% Does this subspace really exist? The work of~\citet{aghajanyan-etal-2021-intrinsic} explores a side of this issue via finding a single subspace for a specific NLP task by projecting $\Delta$ of all model parameters of a PLM with a random linear matrix. However, this still does not answer the problem of the existence of a general subspace shared by various NLP tasks. Moreover, the random linear projections~\citep{li2018measuring,aghajanyan-etal-2021-intrinsic} inevitably introduce redundant task-irrelevant information learned during pre-training into the subspace. In remedy for this, we  of the tuned parameters in a PLM. Once such an intrinsic subspace is found, we assess its effectiveness by investigating how well can we recover the performance for these tasks with previously unseen data and the generalization ability on unseen tasks.

\paragraph{Unifying Different NLP Tasks.}
Although various NLP tasks differ a lot on the surface, there have been long-standing attempts to unify different NLP tasks into the same form~\citep{Sun2021ParadigmSI} and thus handle them with similar techniques, especially after the success of the prompting methods~\citep{Liu2021PretrainPA} to cast various tasks into the form of pre-training tasks of PLMs. The analyses in this paper may help us understand why this can be possible and explore how to better unify different tasks from the perspective of intrinsic task subspace.
\section{Methodology}

We first introduce essential preliminaries for both fine-tuning and prompt tuning in \cref{sec:preliminary}, and then introduce our proposed analysis pipeline \textbf{I}ntrinsic \textbf{P}rompt \textbf{T}uning (\textbf{IPT}) in \cref{sec:IPT}, which consists of two stages: (1) Multi-task Subspace Finding (MSF) and (2) Intrinsic Subspace Tuning (IST). In Figure~\ref{fig:method}, we visualize the paradigms of fine-tuning, prompt tuning and our \ourmodel.

\subsection{Preliminaries}
\label{sec:preliminary}
Assume we are given a series of NLP tasks $\{\mathcal{T}_1, \ldots, \mathcal{T}_{|\mathcal{T}|}\}$ partitioned into training tasks $\mathcal{T}_\text{train}$ and test tasks $\mathcal{T}_\text{test}$. Following~\citet{raffel2020exploring}, without loss of generality, we cast each task $\mathcal{T}_i$ into the unified conditional generation format. Given a training instance $(\mathcal{X}, \mathcal{Y})$ of $\mathcal{T}_i$, where both the input $\mathcal{X}$ and the target $\mathcal{Y}$ consist of a sequence of tokens, i.e., $\mathcal{X} \! = \! \{w_1, \ldots, w_{|\mathcal{X}|}\}$ and $\mathcal{Y} \! = \! \{y_1, \ldots, y_{|\mathcal{Y}|}\}$. Our goal is to learn a mapping function $\mathcal{F}_i: \! \mathcal{X} \! \rightarrow \! \mathcal{Y}$, and the de-facto way is to model $\mathcal{F}_i$ with a PLM $\mathcal{M}$, which first converts the input $\mathcal{X}$ into embeddings $\mathbf{E} \! = \! \{\mathbf{w}_1, \ldots, \mathbf{w}_{|\mathcal{X}|}\} \! \in \! \mathbb{R}^{|\mathcal{X}| \times d}$, where $d$ denotes the input embedding dimension, then encodes $\mathbf{E}$ into hidden representations $\mathbf{H} = \{\mathbf{h}_1, \ldots, \mathbf{h}_{|\mathcal{X}|}\} \in \mathbb{R}^{|\mathcal{X}| \times d}$ and finally decodes $\mathcal{Y}$ conditioning on $\mathbf{H}$. The goal is to optimize the following objective:
\begin{equation}
\begin{aligned}
\mathcal{L}_\text{LM} \! = \! -\frac{1}{|\mathcal{Y}|}\prod\limits_{j=1}^{|\mathcal{Y}|}p(y_j|w_1, ..., w_{|\mathcal{X}|},y_1, ..., y_{j-1}). \nonumber
\end{aligned}
\end{equation}

In traditional fine-tuning, all parameters of $\mathcal{M}$ ($\theta_{\mathcal{M}}$) are tuned during the optimization. Rather, prompt tuning (PT) % Recently, prompt tuning~\cite{lester2021power} springs up as an effective alternative with extensively fewer tunable adaptive parameters. Formally, prompt tuning introduces additional information into the input $\mathcal{X}_i$
prepends some task-specific embeddings (i.e., \textit{soft prompts}) $\mathbf{P}_i \! = \! \{\mathbf{p}_1, \ldots, \mathbf{p}_n\}$ parameterized by $\theta_{P}$ before $\mathbf{E}$, and thus modify the input embeddings into $\mathbf{E}^* = \{\mathbf{p}_1, \dots, \mathbf{p}_n; \mathbf{w}_1, \ldots, \mathbf{w}_{|\mathcal{X}|}\} \in \mathbb{R}^{(n + |\mathcal{X}|) \times d}$. Then we keep $\theta_{\mathcal{M}}$ frozen and only tune $\theta_{P}$ to adapt $\mathcal{M}$ to $\mathcal{T}_{i}$ during PT. The training objective of PT is essentially the same as $\mathcal{L}_\text{LM}$ and denoted as $\mathcal{L}_\text{LM}(\mathbf{P}_i)$.

\subsection{Intrinsic Prompt Tuning}
\label{sec:IPT}
To verify our hypothesis that the adaptations of PLMs to various downstream tasks can be reparameterized as optimization within a unified low-dimensional \textit{intrinsic task subspace}, we propose a two-phase analysis pipeline \ourmodel. The first phase MSF aims to find the intrinsic task subspace with multiple tasks' prompts, which are defined by an auto-encoder consisting of a projection function and a back-projection function. The second phase IST tunes a low-dimensional vector in the subspace and then recovers the vector to soft prompts through the back-projection function. 

\paragraph{Multi-task Subspace Finding.}
% Learning a projection from the prompt subspace to the original space can be formulated as factorizing a task-specific prompt $\mathbf{P} \in \mathbb{R}^{n \times d}$ into $\mathbf{Proj}(\mathbf{d})$, where $\mathbf{Proj}$ can be viewed as $d_I$ bases, each representing a kind of few-shot language skill, and the intrinsic subspace vector $\mathbf{d}_I \in \mathbb{R}^{d_I}$ can be viewed as the coefficients applied on each base. Thus the functionality of a prompt can be explained as continuous compositions of limited few-shot language skills. Specifically, we implement $\mathbf{Proj}$ with a standard two-layer MLP as follows:
We first conduct prompt tuning for each downstream task $\mathcal{T}_i$ and obtain the trained soft prompts $\mathbf{P}_{i} \in \mathbb{R}^{n \times d}$. During MSF, we try to find a satisfactory intrinsic task subspace of a low dimension $d_{I}$ by learning a decomposition for the matrix $\mathbf{P}_{i}$. Inspired by text autoencoders~\citep{bowman-etal-2016-generating}, the decomposition consists of a projection function $\mathbf{Proj}(\cdot)$ to project $\mathbf{P}_{i}$ into the $d_{I}$-dimensional subspace and a back-projection function $\mathbf{Proj}_{b}(\cdot)$ to project the $d_{I}$-dimensional vectors back into soft prompts of $\mathcal{T}_{i}$, and we optimize the reconstruction loss $\mathcal{L}_\text{AE}^i$:
\begin{equation}
\begin{aligned}
\mathbf{P}_i^* &= \mathbf{Proj}_{b}(\mathbf{Proj}(\mathbf{P}_i)), \\
\mathcal{L}_\text{AE}^i &= ||\mathbf{P}_i^* - \mathbf{P}_i||^2_2, \nonumber
\end{aligned}
\end{equation}
where $\mathbf{Proj}(\cdot)$ is implemented with a one-layer feed-forward network and $\mathbf{Proj}_{b}(\cdot)$ is parameterized by a two-layer nonlinear perceptron. %  as follows:

% \begin{equation}
% \begin{aligned}
% \mathbf{Proj}_{b}(\mathbf{d}_i) = \mathbf{W}_2(\tanh(\mathbf{W}_1\mathbf{d}_i + \mathbf{b}_1)) + \mathbf{b}_2, \nonumber
% \end{aligned}
% \end{equation}
% where $\mathbf{W}_1 \in \mathbb{R}^{d_I' \times d_I}$, $\mathbf{b}_1 \in \mathbb{R}^{d_I'}$, $\mathbf{W}_2 \in \mathbb{R}^{n \times d \times d_I'}$ and $\mathbf{b}_2 \in \mathbb{R}^{n \times d}$ are trainable parameters. % In this way, different from~\citet{aghajanyan-etal-2021-intrinsic}, the subspace defined by $\mathbf{Proj}$ and $\mathbf{Proj}_{b}$ will be nonlinear.

Moreover, finding the decomposition of a certain task's prompt $\mathbf{P}_{i}$, which is essentially a matrix, is somewhat trivial. Since the desired intrinsic task subspace should work for broad tasks, we introduce multi-task training and also take the task-oriented language modeling (using the reconstructed soft prompts) losses as objective functions. By jointly optimizing the reconstruction losses and the task-oriented losses, the subspace could gain the ability to reparameterize various task adaptations.
% high-performance soft prompts for the training tasks, so that can be at least the intersection of the low-dimensional subspaces compressing various training tasks.
The overall training objective of MSF is as follows:
\begin{equation}
\begin{aligned}
\mathcal{L}_{\theta_\text{proj}}^{\text{MSF}} = \frac{1}{|\mathcal{T}_\text{train}|}\sum\limits_{i=1}^{|\mathcal{T}_\text{train}|}(\mathcal{L}_\text{LM}(\mathbf{P}_i^*) + \alpha \mathcal{L}_\text{AE}^i), \nonumber
\end{aligned}
\end{equation}
where $\alpha$ denotes the hyper-parameter controlling the ratio between the two losses, and $\theta_\text{proj}$ denotes the parameters of both $\mathbf{Proj}$ and $\mathbf{Proj}_b$. During MSF, we only optimize $\theta_\text{proj}$ while keeping other parameters fixed. By introducing downstream task supervision and nonlinearity, we could find more irredundant and effective subspaces than the random linear subspaces~\citep{aghajanyan-etal-2021-intrinsic}.

\paragraph{Intrinsic Subspace Tuning.}
In this stage, we want to evaluate if the subspace found by MSF is generalizable to previously (1) unseen training data of $\mathcal{T}_\text{train}$ and (2) unseen tasks $\mathcal{T}_\text{test}$. If the answer is yes, we can say that we successfully find the intrinsic task subspace reparameterizing the adaptations of PLMs to various tasks to some extent. Specifically, we only retain $\mathbf{Proj}_{b}$ learned during MSF and keep both $\mathbf{Proj}_{b}$ and $\mathcal{M}$ fixed. Then for each task $\mathcal{T}_i$, instead of conducting vanilla prompt tuning, we tune only $d_I$ (bottleneck dimension) free parameters ($\theta_{d}$) in the found subspace, which form a randomly initialized \textit{intrinsic vector} $\mathbf{V}_{i}\in \mathbb{R}^{d_I}$, and project them into soft prompts with the fixed $\mathbf{Proj}_{b}$. The objective function for training a specific task $\mathcal{T}_i$ could be formulated as:
\begin{equation}
\begin{aligned}
\mathcal{L}_{\theta_{d}}^{\text{IST}} = \mathcal{L}_\text{LM}(\mathbf{Proj}_{b}(\mathbf{V}_{i})). \nonumber
\end{aligned}
\end{equation}
%n this way, we obtain $d_I$ bases, each representing a kind of few-shot NLP skill. In $\mathbf{IST}$, we aim to learn a continuous composition of these bases. Specifically, we keep both $\mathbf{Proj}$ and $\mathcal{M}$ fixed, and optimize $\mathcal{L}_\text{LM}$ for each task individually tuning only $d_I$ parameters in the prompt subspace. $\mathbf{IST}$ is expected to perform well if most language skills required for solving few-shot NLP tasks are covered in the parameters of both $\mathbf{Proj}$ and $\mathcal{M}$.
\section{Experiment and Analysis}
\label{sec:exp}
In this section, we first describe the experimental settings in~\cref{sec:setting}, including the tasks and corresponding datasets, evaluation pipeline, evaluation metrics and training details. Then we introduce the experimental results and analyses in~\cref{exp:main} and~\cref{exp:analysis}. We provide notation descriptions in \cref{sec:notation}.

\subsection{Experimental Settings}
\label{sec:setting}
\paragraph{Tasks and Datasets.}
To cover broad NLP tasks, we randomly choose $120$ few-shot NLP tasks ($\mathcal{T}_{\text{all}}$, see \cref{sec:ontology} for details) from \textit{CrossFit Gym}~\citep{ye2021crossfit}, including text classification, question answering, conditional generation, etc.
To evaluate the generalization ability of \ourmodel, we randomly split the overall task set $\mathcal{T}_{\text{all}}$ into training tasks $\mathcal{T}_\text{train}$ and test tasks $\mathcal{T}_\text{test}$. We adopt three task splits as listed in Table~\ref{tab:task_split} to investigate the influence of task types.
% (2) the performance gaps between soft prompting and fine-tuning are narrower under the few-shot settings, which will make the results more representative. 
Each task $\mathcal{T}_i \in \mathcal{T}_\text{all}$ consists of a tuple of ($\mathcal{D}^i_\text{train}$, $\mathcal{D}^i_\text{dev}$, $\mathcal{D}^i_\text{test}$), and the sizes of $\mathcal{D}^i_\text{train}$ and $\mathcal{D}^i_\text{dev}$ are both set to $K$ for the few-shot setting. For classification and regression tasks, $K = 16$; while for other categories, $K = 32$. The few-shot setting ensures the data scales of tasks are balanced so that the subspace found by MSF will not be easily biased towards data-rich tasks.
% For instance, the input of a multi-choice QA task is formulated as: ``\texttt{Question:} \texttt{<question>} \texttt{Context:} \texttt{<context>} \texttt{Candidates:} \texttt{<candidates>}'', and the PLM is expected to generate the correct answer span from \texttt{<candidates>}.

\begin{table}[!tbp]
  \centering
  \small
  \setlength{\tabcolsep}{2.8pt}
  {

    \begin{tabular}{ccc}
    \toprule
    \multicolumn{1}{c}{\multirow{1}[1]{*}{\textbf{Shorthand}}} & \multirow{1}[1]{*}{$\mathcal{T}_\text{train}$} & \multicolumn{1}{c}{\multirow{1}[1]{*}{$\mathcal{T}_\text{test}$}} \\
    \midrule
    \textit{random} & $100$ random  &  $20$ random  \\
    \textit{non-cls} &  $35$ non-cls.     &  $42$ non-cls.($\mathcal{T}_\text{test}^\text{in}$) / $43$ cls.($\mathcal{T}_\text{test}^\text{out}$)   \\
    \textit{cls}   &  $35$ cls.     &  $8$ cls.($\mathcal{T}_\text{test}^\text{in}$) / $77$ non-cls.($\mathcal{T}_\text{test}^\text{out}$) \\
    \bottomrule
    \end{tabular}
  }
  \caption{The overall $120$ tasks $\mathcal{T}_{\text{all}}$ consist of $43$ classification tasks (cls.) and $77$ non-classification tasks (non-cls.). Three task splits are evaluated, including \textit{random}, \textit{non-cls} and \textit{cls}, with details listed above, e.g., for \textit{non-cls} partition, $35$ non-cls. are chosen as $\mathcal{T}_\text{train}$ and $42$ non-cls. / $43$ cls. are chosen as $\mathcal{T}_\text{test}^\text{in}$ / $\mathcal{T}_\text{test}^\text{out}$, respectively.  }
  \label{tab:task_split}%
\end{table}%

\paragraph{Evaluation Pipeline: MSF.}
During MSF, we first conduct prompt tuning for each task and obtain the soft prompts. Then we train an auto-encoder on $\mathcal{T}_\text{train}$, and (1) evaluate the reconstructed prompts on $\mathcal{T}_\text{train}$ (denoted as $\mathcal{T}_\text{train}(\text{MSF})$) to see how much performance we could retain after reconstruction from a $d_I$-dimensional subspace. This performance provides an empirical upper bound for the generalization to unseen data and tasks in our setting. (2) We also directly reconstruct the soft prompts of $\mathcal{T}_\text{test}$ with the learned auto-encoder and test their performance ($\mathcal{T}_\text{test}(\text{MSF})$) to see the auto-encoder's reconstruction ability for unseen soft prompts.

\paragraph{Evaluation Pipeline: IST.}
During IST, we investigate whether adaptations to various tasks can be reparameterized into the found subspace.

We first carry out experiments on $\mathcal{T}_\text{train}$ using exactly \textbf{the same} $\mathcal{D}^i_\text{train}$ / $\mathcal{D}^i_\text{dev}$ utilized in MSF training, and get the result $\mathcal{T}_\text{train}^\text{same}(\text{IST})$.
% can be roughly viewed as the upper bound for subsequent experiments
Then we evaluate the generalization ability of \ourmodel with two challenges: (1) \textit{unseen-data challenge} and (2) \textit{unseen-task challenge}.

$\bullet$ For the \textit{unseen-data challenge}, we sample \textbf{different} $\mathcal{D}^i_\text{train}$ / $\mathcal{D}^i_\text{dev}$ for $\mathcal{T}_\text{train}$ while keeping test data the same. Then we conduct IST with the new data and test its performance on $\mathcal{T}_\text{train}$, which is denoted as $\mathcal{T}_\text{train}^\text{diff}(\text{IST})$. This challenge evaluates whether the learned subspace can also reparameterize optimization on unseen data, which leads to different optimization trajectories than the seen data. %Note $\mathcal{D}^{i'}_\text{train}$ / $\mathcal{D}^{i'}_\text{dev}$ and $\mathcal{D}^{i}_\text{train}$ / $\mathcal{D}^{i}_\text{dev}$ conform with the i.i.d. assumption. The \textit{i.i.d. challenge} is designed to test whether the learned $\mathbf{Proj}$ is indeed compressing the universal few-shot language skills for different tasks, instead of over-fitting some meaningless data point clues of $\mathcal{D}^{i}_\text{train}$/$\mathcal{D}^{i}_\text{dev}$.

$\bullet$ For the \textit{unseen-task challenge}, we evaluate the soft prompts obtained by IST on $\mathcal{T}_\text{test}$, which are tasks unseen during MSF. We aim to investigate how well can optimization in the found subspace recover PLM adaptations of unseen tasks, which will provide evidence for our hypothesis that the reparameterization subspaces for different task adaptations are not orthogonal. In the task splits in Table~\ref{tab:task_split}, for the \textit{random} split, the results of this challenge are denoted as $\mathcal{T}_\text{test}(\text{IST})$; for the \textit{non-cls} and \textit{cls} splits, we have two test sets with different task types and the corresponding results are denoted as $\mathcal{T}_\text{test}^\text{in}(\text{IST})$ and $\mathcal{T}_\text{test}^\text{out}(\text{IST})$, respectively.

%we take a step further and evaluate on $\mathcal{T}_\text{test}$. Although it may require additional language skills to well handle $\mathcal{T}_\text{test}$, we hypothesize that most of the language skills are shared by both $\mathcal{T}_\text{train}$ and $\mathcal{T}_\text{test}$, making it possible for cross-task generalization. Ideally, \ourmodel should perform better on \textit{cross-task challenge} if more language skills are properly depicted and retained during MSF.

\begin{figure*}[t!]
\centering
\includegraphics[width=0.95\textwidth]{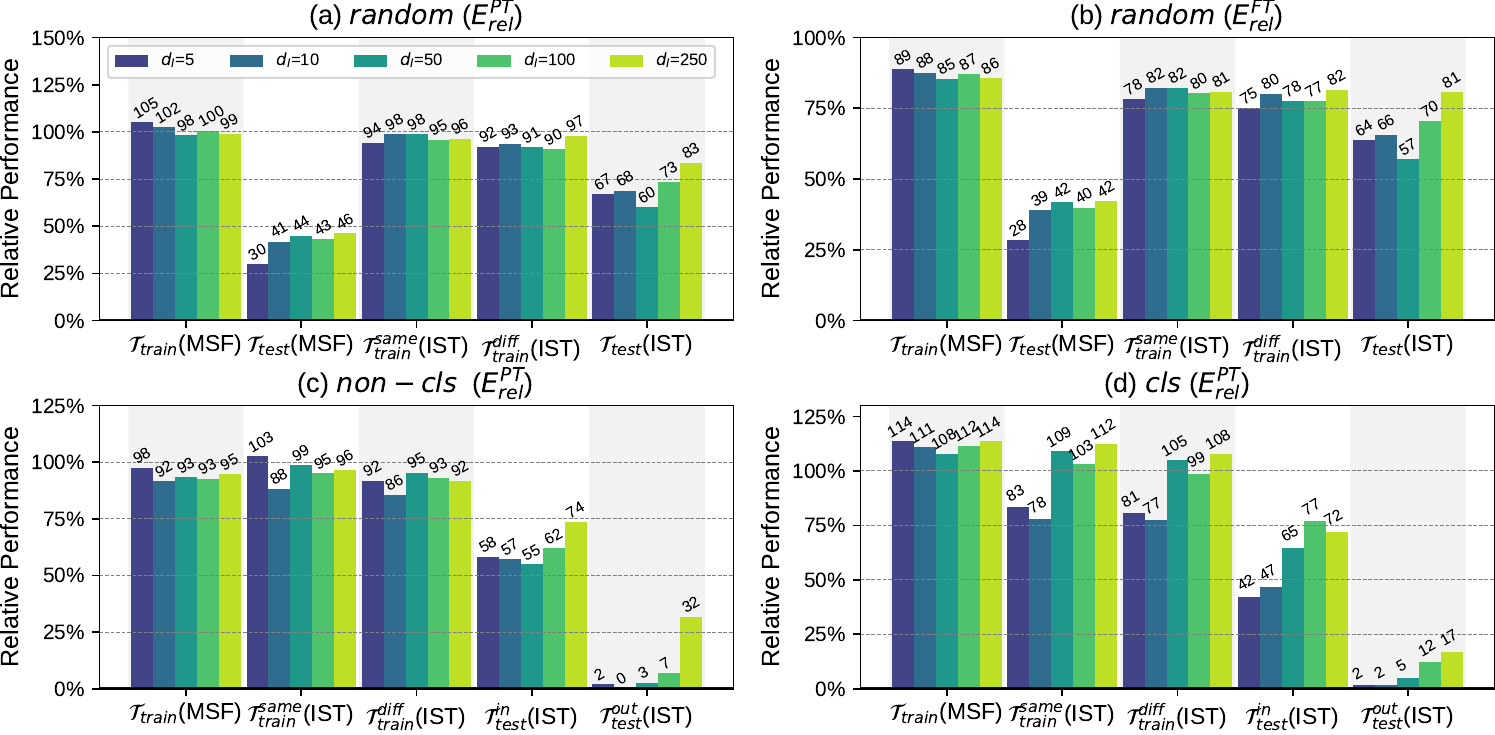}
\caption{Relative performance of \ourmodel at different dimension $d_{I}$ on three task splits (\textit{random}, \textit{non-cls} and \textit{cls}). We report the relative performance of \ourmodel compared with both prompt tuning ($E_\text{rel}^\text{PT}$) and fine-tuning ($E_\text{rel}^\text{FT}$).
}
\label{fig:relative}
\end{figure*}

\paragraph{Evaluation Metrics.}
Since different tasks have distinct evaluation protocols (e.g., F1 score for discriminative tasks and BLEU for generative tasks typically), following \citet{ye2021crossfit}, we mainly choose average relative performance ($E_\text{rel}$) as the evaluation metric, instead of absolute performance (also reported in \cref{sec:app_abs} for reference). Specifically, let $\mathcal{T} \! = \! \{\mathcal{T}_1, ..., \mathcal{T}_{|\mathcal{T}|}\}$ be the evaluated tasks and $E_{\mathcal{T}_i}$ denotes the test score of $\mathcal{T}_i$ for \ourmodel (either in MSF or IST), $E_\text{rel}^* = \frac{1}{|\mathcal{T}|}\sum_{\mathcal{T}_i \in \mathcal{T}}\frac{E_{\mathcal{T}_i}}{E^*_{\mathcal{T}_i}}$, where $E^*_{\mathcal{T}_i}$ denotes the performance of $T_i$ using either prompt tuning ($E_{\mathcal{T}_i}^\text{PT}$) or fine-tuning ($E_{\mathcal{T}_i}^\text{FT}$). $E^{\text{PT}}_{\text{rel}}$ / $E^{\text{FT}}_{\text{rel}}$ denotes our \ourmodel's relative performance to prompt tuning / fine-tuning.

\paragraph{Training Details.}
We use $\text{BART}_{\texttt{BASE}}$~\citep{lewis-etal-2020-bart} for the experiments in the main paper, and unify all tasks into the same sequence-to-sequence format. We also test $\text{BART}_{\texttt{LARGE}}$ in \cref{sec:bartlarge}. For the prompt tuning / fine-tuning baseline, we perform grid search on the combination of a series of learning rates and batch sizes and choose the best checkpoint using $\mathcal{D}_\text{dev}$. We set the number of soft prompts to be $100$ for all tasks and randomly initialize them. For \ourmodel, we examine the dimension $d_I$ of $\{5, 10, 50, 100, 250\}$. %We set the batch size to $256$/$64$ and the max step to $200,000$/$100,000$ for experiments in \ref{exp:main}, respectively.
Note that for fine-tuning / prompt tuning, $139$M / $76,800$ parameters are tuned, while \ourmodel only tunes $d_I$ free parameters. More details are left in \cref{sec:training_detail}.

\subsection{Main Results}
\label{exp:main}

Based on the experimental results shown in Figure~\ref{fig:relative}, we study the following questions:

\paragraph{Q1. Do PLMs really reparameterize various task adaptations into a universal task subspace?}
From the results in Figure~\ref{fig:relative} (a), we observe that: (1) for the \textit{unseen-data challenge} ($\mathcal{T}_\text{train}^\text{diff}(\text{IST})$), IST on unseen i.i.d. data could recover more than $90\%$ of the full prompt tuning performance of the $100$ training tasks; (2) for the \textit{unseen-task challenge} ($\mathcal{T}_\text{test}(\text{IST})$), we can also achieve $83\%$ performance by only tuning $250$ parameters. From these results, we can say that the low-dimensional reparameterizations in the subspaces found by MSF successfully recover the PLM adaptations of $\mathcal{T}_\text{train}$ and can also generalize to unseen tasks. Thus non-trivial performance can be achieved by only tuning a few free parameters in these subspaces. This strongly supports our hypothesis that PLMs reparameterize various task adaptations into the same low-dimensional subspace, or at least the low-dimensional reparameterization subspaces for various task adaptations~\citep{aghajanyan-etal-2021-intrinsic} should have a substantial intersection, otherwise the subspace found with $\mathcal{T}_\text{train}$ in MSF would be almost impossible to also work for $\mathcal{T}_\text{test}$.

%Moreover, in the \textit{random} split, we can see that $\mathcal{T}_\text{train}^\text{diff}(\text{IST})$ and $\mathcal{T}_\text{test}(\text{IST})$ are always on the same level when $d_I \geq 5$, indicating that the dimension for the \textit{intrinsic task subspace} should be around $5$, which is extremely low for a PLM with hundreds of millions of parameters.

\paragraph{Q2. What limits \ourmodel?}
Although strong evidence is observed, the effectiveness of IPT could still be improved, especially at low dimensions. From the results in Figure~\ref{fig:relative}~(a) and~(b), we discuss what factors may limit the effectiveness and provide insights for improving the analysis pipeline.

(1) \textbf{Reconstruction ability of the auto-encoder.} The performance on $\mathcal{T}_\text{train}$ when we directly reconstruct soft prompts using the auto-encoder of MSF ($\mathcal{T}_\text{train}(\text{MSF})$) can be even better than vanilla prompt tuning (PT). This is because MSF explicitly enforces multi-task knowledge sharing within a unified subspace. Such knowledge sharing equips the subspace with better representational capabilities, making the subspace more universal. Thus the performance of MSF could even exceed conducting vanilla PT for each individual task. However, from the comparisons between $\mathcal{T}_\text{train}(\text{MSF})$ and $\mathcal{T}_\text{test}(\text{MSF})$, we can see that directly reconstructing soft prompts of unseen tasks performs poorly. It indicates that the reconstruction ability of the auto-encoders trained in MSF cannot generalize well to unseen soft prompts, which will limit \ourmodel to some extent. This may come from the limited representation ability of the networks used to parameterize $\mathbf{Proj}(\cdot)$ and $\mathbf{Proj}_{b}(\cdot)$. Nevertheless, IST could find better solutions ($\mathcal{T}_\text{test}(\text{IST})$) than MSF reconstructed prompts ($\mathcal{T}_\text{test}(\text{MSF})$) with task-specific supervisions on $\mathcal{T}_{\text{test}}$.

(2) \textbf{Optimization in IST.} The performance of $\mathcal{T}_\text{train}(\text{MSF})$ is always near $100\%$, which demonstrates that there exists good enough solutions for $\mathcal{T}_{\text{train}}$ in the found subspace. However, even using exactly the same training data, IST cannot find these good solutions (the gap between $\mathcal{T}_\text{train}(\text{MSF})$ and $\mathcal{T}_\text{train}^\text{same}(\text{IST})$), which shows that the adopted optimization algorithm may limit IST performance. We also observe that with $d_I$ increasing, the recovering performance of IST generally becomes better, which is because a higher dimension brings larger representational capacities.
% is aligned with \citet{aghajanyan-etal-2021-intrinsic}.

(3) \textbf{Adaptive parameters.} Comparing the results in Figure~\ref{fig:relative} (a) and (b), we observe that the relative performance of fine-tuning ($E_{\text{rel}}^\text{FT}$) is always poorer than that of PT ($E_{\text{rel}}^\text{PT}$). Since both of them have the same numerator, the above phenomenon is because PT is slightly inferior to fine-tuning under the few-shot setting. Since the performance of \ourmodel is bounded by PT, ideally, $E_{\text{rel}}^\text{FT}$ could be improved by designing better PT algorithms or selecting more appropriate adaptive parameters.

\paragraph{Q3. What is the influence of task types?}
We first divide the studied tasks into \textit{cls} (classification), which are discriminative tasks and \textit{non-cls} (non-classification), which tend to be generative tasks. From the results in Figure~\ref{fig:relative} (c)-(d), we find that there exists a huge generalization gap between the two coarse-grained task types. When using only one kind of tasks during MSF, the found subspaces work well for the same kind of tasks ($\mathcal{T}_\text{test}^\text{in}(\text{IST})$) but generalize poorly to the other kind of tasks ($\mathcal{T}_\text{test}^\text{out}(\text{IST})$). This shows that the found subspace is severely biased by the training task types. We further conduct more fine-grained analyses on task types to fathom their influence in \cref{sec:task_type}.
\subsection{Analyses and Properties}
\label{exp:analysis}

\paragraph{Comparison of Subspace-finding Methods.}
\begin{figure}[!t]
\centering
\includegraphics[width=0.45\textwidth]{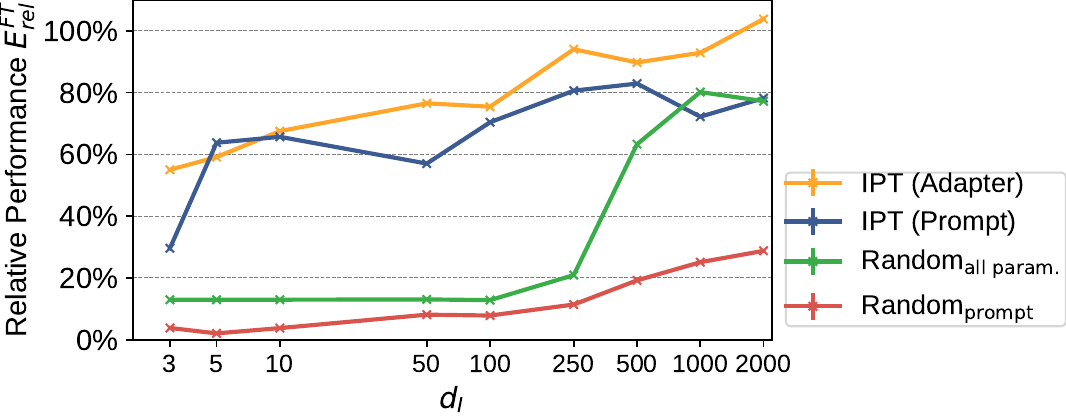}
\caption{Comparisons ($E_\text{rel}^\text{FT}$) between \ourmodel and other subspace-finding methods on $\mathcal{T}_{\mathrm{test}}$ of \textit{random} task split.}
\label{fig:vs_random}
\end{figure}

In Figure~\ref{fig:vs_random}, we compare the relative fine-tuning performance ($E_\text{rel}^\text{FT}$) of subspaces found by \ourmodel with various subspace-finding methods on the \textit{unseen-task} challenge under \textit{random} split\footnote{Results on $\mathcal{T}_{\mathrm{train}}$ are shown in appendix~\ref{sec:random_training}}. (1) First, we investigate the randomly generated subspaces of \citet{aghajanyan-etal-2021-intrinsic}: the first trial Random$_{\mathrm{prompt}}$ conducts IST within random subspaces of \textit{soft prompts}. The subspaces are defined by randomly initialized auto-encoders of the same architecture as MSF; the second trial Random$_{\mathrm{all~param.}}$ conducts IST within random subspaces of \textit{all PLM parameters}, which are generated by the efficient Fastfood transform~\citep{le2013fastfood}. We can see that the random subspace-finding methods can also find effective unified reparameterization subspaces at large dimensionality, which supports our universal reparameterization subspace hypothesis. In addition, \ourmodel performs much better using much fewer dimensions, which indicates the effectiveness of MSF to exclude redundant task-irrelevant information and find compact subspaces. (2) Then we explore another parameter-efficient tuning method, Adapter~\citep{Houlsby2019Adapter}, as the backbone of our method, i.e., we conduct IPT pipeline using the adapter parameters\footnote{The implementation details are shown in appendix~\ref{sec:training_detail_adapter}.} instead of soft prompts. We observe that this method (IPT (Adapter)) performs consistently better than the original \ourmodel using soft prompts (IPT (Prompt)) and can even outperform fine-tuning at $2000$ dimensionality, which may be due to the better performance of adapter than prompt tuning~\citep{Hu2021LoRALA}. This further shows that \ourmodel is agnostic to the specific tuning method and provides stronger empirical evidence for our research hypothesis. %Previous works~\citep{li2018measuring,aghajanyan-etal-2021-intrinsic} adopt randomly generated subspaces and avoid computation in subspace finding. While in this work, we introduce supervisions from diverse tasks to find the universal low-dimensional intrinsic task subspaces. To verify the effectiveness and necessity of task-specific supervisions in MSF, we compare \ourmodel with conducting IST in randomly generated subspaces, which are defined by randomly initialized auto-encoders of the same architecture with the ones used in MSF. We compare them under the \textit{random} task split. For \ourmodel, we report the \textit{unseen-data} ($\mathcal{T}_\text{train}^{\text{diff}}$(\text{IST})) and \textit{unseen-task} ($\mathcal{T}_\text{test}$(\text{IST})) performance. For random subspaces, we also report their performance on $\mathcal{T}_{\text{train}}$ (denoted as $\mathcal{T}_{\text{train}}$(Random)) and $\mathcal{T}_{\text{test}}$ ($\mathcal{T}_{\text{test}}$(Random)), respectively. The results are shown in Figure~\ref{fig:vs_random}, from which we can see that \ourmodel could perform much better than random subspaces using much fewer dimensions, which indicates the effectiveness of MSF to exclude redundant task-irrelevant information and find compact reparameterization subspaces. % (2) when $d_I$ grows larger, non-trivial performance can also be obtained within random subspaces, which supports our hypothesis that various task adaptations can be reparameterized into a unified subspace conditioning on PLMs.

\paragraph{Impacts of the Data Scale.}

\begin{figure}[!t]
\centering
\includegraphics[width=0.46\textwidth]{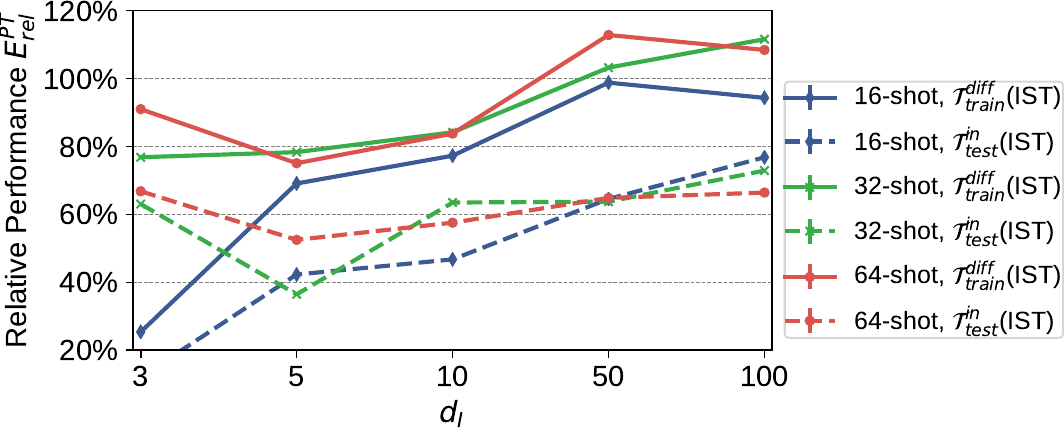}
\caption{Impacts of the data scale.}
\label{fig:shots}
\end{figure}

Although we adopt the few-shot setup to control the influence of data amount in this paper, it is also interesting to investigate \ourmodel's ability given more training data. Here we take an initial trial using the task split \textit{cls} by doubling / quadrupling the number of data shots $K$ of both seen and unseen tasks (from $16$ to $32$ / $64$), and investigate the performance of \ourmodel under the \textit{unseen-data} ($\mathcal{T}_\text{train}^{\text{diff}}$(\text{IST})) and \textit{unseen-task} ($\mathcal{T}_\text{test}^{\text{in}}$(\text{IST})) challenges. Note that with different number of data points, the PT performance (denominator of $E_\text{rel}^\text{PT}$) is also different. The results are shown in Figure~\ref{fig:shots}, from which we observe that when the data scale grows, the performance of \ourmodel generally becomes better, especially at low dimensions. This shows that the subspace approximated with more data is more universal. Hence we encourage future work to explore how strong the performance of \ourmodel on data-rich scenarios will be.
%: (1) the gap between $\mathcal{T}_\text{train}$(\text{MSF}) and $\mathcal{T}_\text{train}^{\text{diff}}$(\text{IST}) tends to be narrower, which demonstrates that involving more supervision makes the subspace found by MSF more robust to data shift; (2) the performance of \ourmodel on unseen tasks ($\mathcal{T}_\text{test}^{\text{in}}$(\text{IST})) generally become better, which shows the subspaces found with more data are generally more universal. Hence we believe it is interesting to explore how strong the subspace found by \ourmodel on data-rich scenarios will be. 
% We also find that when increasing $d_I$, $\mathcal{T}_\text{train}^\text{same}$ for $2K$-shot grows very close to $100\%$ but never exceeds it, indicating that when more data is available, jointly reconstructing prompts from muliple tasks does not bring additional benefits as observed in the $K$-shot setting. 

\paragraph{Impacts of the Number of Training Tasks.}

\begin{figure}[!t]
\centering
\includegraphics[width=0.45\textwidth]{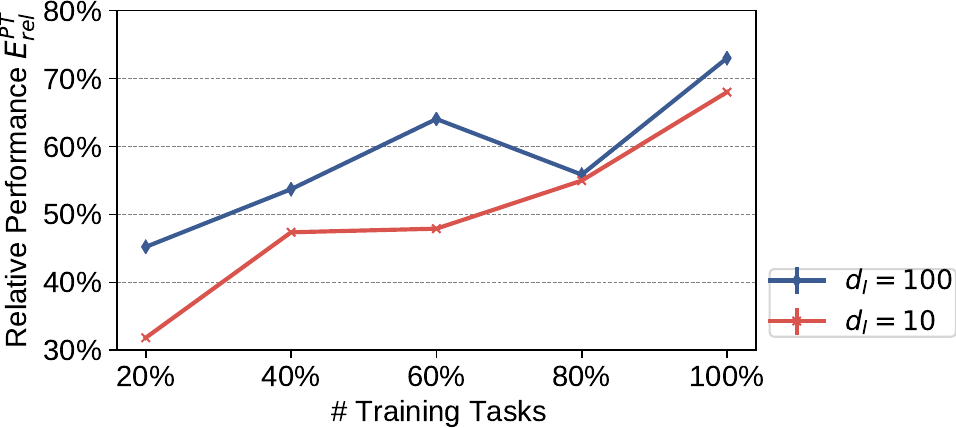}
\caption{Impacts of the number of training tasks.
}
\label{fig:task_num}
\end{figure}

During MSF, the auto-encoder is optimized to reparameterize the adaptive parameters of various training tasks. Ideally, the coverage of $\mathcal{T}_\text{train}$ would significantly impact the generalization ability of \ourmodel on unseen tasks $\mathcal{T}_\text{test}$. To demonstrate this, we randomly sample $\{20\%,40\%,60\%,80\%\}$ tasks from $\mathcal{T}_\text{train}$ of the \textit{random} task split to train the auto-encoder, then evaluate \ourmodel ($d_I=\{10,100\}$) on original $\mathcal{T}_\text{test}$ with the \textit{unseen-task} challenge. From the results visualized in Figure~\ref{fig:task_num}, we observe that with the number of training tasks growing, the generalization ability of the found intrinsic task subspace generally improves. This reflects that increasing the coverage and diversity of seen tasks could help \ourmodel find more universal subspaces.

\paragraph{Visualization of the Found Intrinsic Subspace.}

\begin{figure}[!t]
\centering
\includegraphics[width=0.45\textwidth]{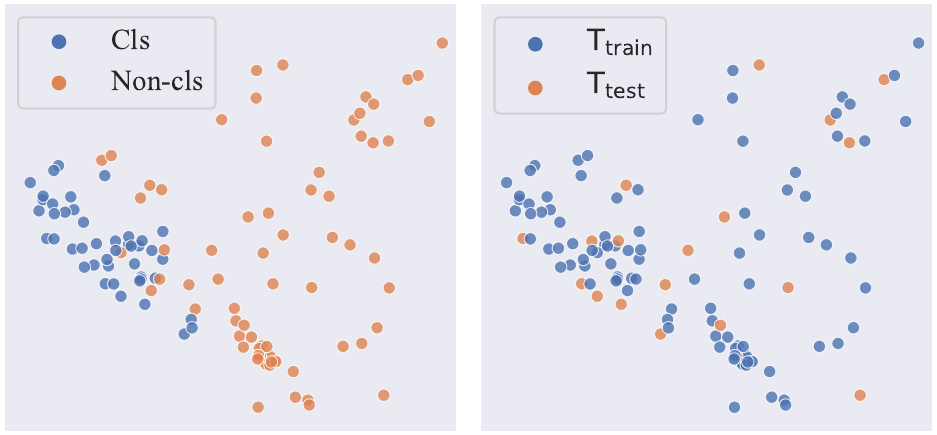}
\caption{PCA plots of the intrinsic vectors learned during IST. We label points with different colors to represent their corresponding categories. Specifically, we show the clusters of (1) classification and non-classification tasks (left) and (2) $\mathcal{T}_\text{train}$ and $\mathcal{T}_\text{test}$ (right). Without loss of generality, we choose the task split of \textit{random} and $d_I = 100$.}
\label{fig:visualization}
\end{figure}

We visualize the intrinsic vectors $\mathbf{V}_{i}$ (the free parameters learned during IST in the found subspace) using PCA in Figure~\ref{fig:visualization}, from which we observe that: (1) there exists a clear dividing line between the clusters of classification tasks and non-classification tasks, indicating that they are highly distinct. This also explains why the subspace learned on one cluster generalizes poorly to the other cluster; (2) the points of unseen tasks $\mathcal{T}_\text{test}$ are mixed with those of $\mathcal{T}_\text{train}$, which demonstrates that the found subspaces universally reparameterize various tasks so that \ourmodel can generalize well to unseen tasks. We also visualize the clusters of fine-grained categories of QA and text classification tasks in \cref{sec:visual_2}. We contend that the learned intrinsic vectors could be viewed as low-dimensional task representations, helping us analyze the similarity and differences of various NLP tasks.% (3) It is also observed from (c) and (d) that the points belonging to the same category exhibit a compact cluster. We argue that the learned intrinsic vectors could be viewed as low-dimensional task representations, helping analyze the similarity and difference for various NLP tasks.

\paragraph{Improving Prompt Tuning Stability with \ourmodel.}

In Table \ref{tab:variance}, we show the mean standard deviations (\texttt{std}) of test scores for $120$ few-shot tasks over $10$ runs comparing \ourmodel ($d_I$ = $10$), fine-tuning and prompt tuning (PT). We observe that PT is the most unstable strategy with the highest \texttt{std}, while fine-tuning is far more stable. The instability of PT may influence its practical uses. Intuitively, \ourmodel only tunes a few free parameters, which will conduce to improving the stability, and \ourmodel surely becomes the most stable method in Table~\ref{tab:variance}. % We further show in \cref{sec:combine} that IPT and vanilla PT could be combined in a two-stage manner to improve both stability and performance. % However, as shown in Figure~\ref{fig:relative}, \ourmodel suffers a performance drop for unseen tasks.
% To make the stability advantage brought by \ourmodel practical

Furthermore, we propose to use the solutions found by \ourmodel as the initialization for the vanilla PT. Specifically, we conduct vanilla PT of split \textit{random} on $\mathcal{T}_\text{test}$ choosing $d_I = 10$ and initialize the soft prompts by back-projecting the IST solutions in the subspace. Other details are kept the same as the PT baseline. We observe that the \texttt{std} achieved in this way is significantly lower than the vanilla PT ($1.65$ v.s. $4.19$) and we can achieve $103.4\%$ of $E_\text{rel}^\text{PT}$, i.e., the performance could also be improved. This indicates that \ourmodel and vanilla PT can be combined in a two-stage manner to improve the training stability and achieve satisfying performance.
% to improve both the training stability and performance. In other words, \ourmodel could find an initialization beneficial to stable training. % This experiment also demonstrates that although our \ourmodel pipeline mainly works as an analytical framework in this paper, it can also bring practical benefits. We will explore more practical uses of \ourmodel in the future.

% To make the stability advantage brought by \ourmodel practical, we propose to use the solutions found by \ourmodel as the initialization for the vanilla prompt tuning. Specifically, we continue the experiments of partition \textit{random} on $\mathcal{T}_\text{test}$ choosing $d_I = 10$ and initialize the soft prompts by back-projecting the found solutions in the subspace during IST. Other details are kept the same compared with prompt tuning baseline. We observe that the standard variance achieved in this way is significantly lower than the vanilla prompt tuning ($1.65$ v.s. $4.19$) while in this way we can achieve $103.4\%$ of $E_\text{rel}^\text{PT}$, i.e., the performance could also be improved from $59\%$ (IST). This indicates that both \ourmodel and prompt tuning could be further combined in a two-stage manner. This experiment also demonstrates that although our \ourmodel pipeline mainly works as an analytical framework in this paper, it can also bring practical benefits. We will explore more practical use of \ourmodel in future.

\begin{table}[!tbp]
  \small
  \centering
    \begin{tabular}{lccc}
    \toprule
    \textbf{Method}     & $\mathcal{T}_\text{train}$  & $\mathcal{T}_\text{test}$ & $\mathcal{T}_{\text{all}}$ \\
    \midrule
  Fine-tuning    & $2.16$     & $2.40$  & $2.20$ \\
  Prompt Tuning    & $3.06$     & $4.19$  & $3.25$ \\
  \ourmodel    & $\textbf{1.12}$     & $\textbf{0.73}$  & $\textbf{1.06}$ \\
    \bottomrule
    \end{tabular}%
  \caption{Standard deviations (\texttt{std}) of test scores over multiple runs. $d_I$ of \ourmodel is chosen to be $10$.}
  \label{tab:variance}%
\end{table}%
\section{Conclusion and Discussion}
% \paragraph{Could few-shot NLP tasks be reparameterized into a unified subspace?}
We study the hypothesis that PLM adaptations to various tasks can be reparameterized as optimizations within a \textbf{unified} low-dimensional \textit{intrinsic task subspace}. We develop an analysis tool \ourmodel. It first finds a subspace by jointly decomposing the adaptive parameters of multiple tasks and then tunes parameters within the subspace for unseen data and tasks. Experiments show the found subspaces contain good solutions for PLM adaptations, which is strong evidence for our hypothesis.

% The achieved performance of \ourmodel is still far from perfect. Although it may come from the inadequacy of current subspace-finding methods and optimization algorithms as mentioned in our analyses, based on current results, we cannot directly conclude that the hypothesis is true. Nevertheless, at least we have found promising empirical results showing that the low-dimensional reparameterization subspaces of various tasks have a substantial \textbf{intersection}, which MSF is designed to find. 

\textbf{Relation to the scaling law.} Recently, researchers have found that larger PLMs tend to be more sample-efficient~\citep{Kaplan2020ScalingLF}, parameter-efficient~\citep{lester2021power} and cross-task generalizable~\citep{Wei2021FLAN}. Our hypothesis may help us understand this phenomenon: the adaptations of larger PLMs can be better reparameterized into a unified subspace so that the cross-task generalization will be easier, and \citet{aghajanyan-etal-2021-intrinsic} show larger PLMs have lower reparameterization dimensions, hence they should need fewer data and tunable parameters. This also implies that the characteristics of intrinsic task subspaces could be used to examine how well a PLM is trained. 

\textbf{Utilize and manipulate intrinsic vectors.} The intrinsic vectors obtained during IST depict the adaptations to different tasks and it is worthwhile to explore whether we can (1) utilize them to find the relations among different tasks, and (2) manipulate these vectors to achieve cross-task generalization. We also encourage future works to explore more methods to tune PLMs within low-dimensional intrinsic task subspaces, which may have some practical benefits such as avoiding over-parameterization and being more environmentally friendly with fewer tunable parameters.

% \paragraph{What's next?} In future, we will explore (1) how to improve \ourmodel to find stronger evidence for our hypothesis, (2) whether the \textbf{union} of reparameterization subspaces for various tasks is also low-dimensional. We also encourage further explorations based on our hypothesis, such as (1) understanding the scaling law of PLMs, (2) how to utilize and manipulate intrinsic vectors, and (3) how to better tune PLMs in the intrinsic task subspaces. % We leave the detailed discussions in \cref{sec:add_discuss}.

%In this paper, we study the hypothesis that the adaptations of PLMs to various tasks can be reparameterized as optimizations of a few free parameters in the same low-dimensional \textit{intrinsic task subspace}. We develop an analysis pipeline called \ourmodel, which first finds a subspace by jointly compressing the adaptive parameters of multiple tasks and then tune parameters only in the subspace for unseen data and tasks. The non-trivial performances achieved by \ourmodel provide positive evidence for the hypothesis. 
%We also discuss the factors influencing the results and the potential practical use of \ourmodel.

\bibliography{anthology,custom}
\bibliographystyle{acl_natbib}

\clearpage
\appendix
\section*{Appendices}
\label{apdx}

\section{Additional Experiments}

\subsection{Absolute Performance}
\label{sec:app_abs}

In the experiments, we mainly report the relative performance ($E_\text{rel}$). For reference, we also report the average absolute performance ($E_\text{abs}$) in this section. Let $E_{\mathcal{T}_i}$ denote the test score of $\mathcal{T}_i$ for \ourmodel, $E_\text{abs} = \frac{1}{|\mathcal{T}|}\sum\limits_{\mathcal{T}_i \in \mathcal{T}}E_{\mathcal{T}_i}$. The $E_\text{abs}$ of $\text{BART}_{\texttt{BASE}}$ for prompt tuning and fine-tuning are shown in Table~\ref{tab:abs_pt_ft}, and the $E_\text{abs}$ of \ourmodel on three task splits are shown in Table~\ref{tab:abs_random}, Table~\ref{tab:abs_non_cls} and Table~\ref{tab:abs_cls}, respectively.

\subsection{Relative Performance to Fine-tuning}

\begin{figure}[!th]
\centering
\includegraphics[width=0.47\textwidth]{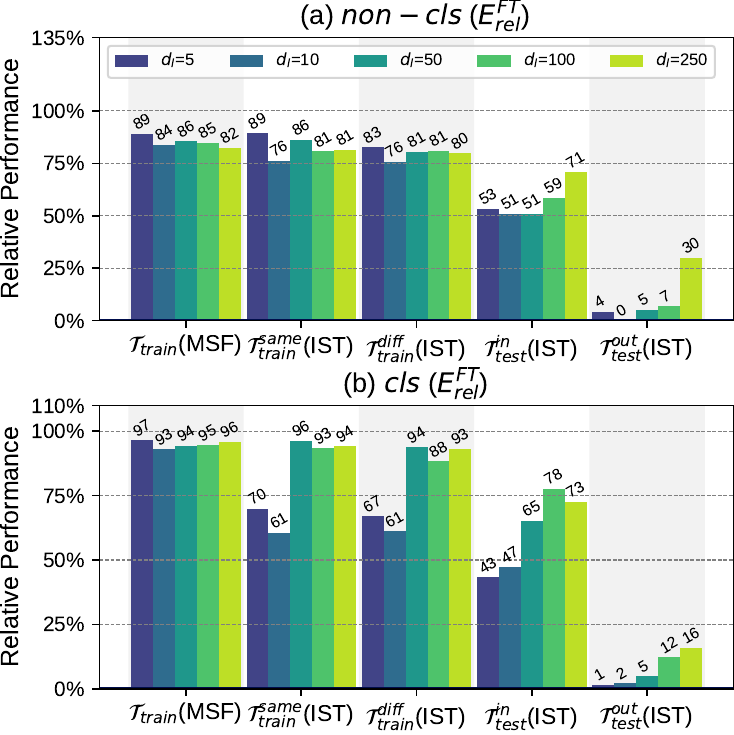}
\caption{Relative performance of \ourmodel at different dimension $d_{I}$ on \textit{non-cls} and \textit{cls} splits, compared with fine-tuning.
}
\label{fig:relativeFT}
\end{figure}

In the experiments, we report the relative performance to prompt tuning as the main evaluation metric except in Figure~\ref{fig:relative} (b), which reports the relative performance to fine-tuning on the \textit{random} split for analyses. In this section, we additionally report $E_{\text{rel}}^{\text{FT}}$ on \textit{non-cls} and \textit{cls} splits in Figure~\ref{fig:relativeFT} for reference, where we can see the general conclusions are consistent with our analyses in \cref{exp:main}.% and the gap between prompt tuning and fine-tuning exists in both two types of tasks.

\subsection{$\text{BART}_{\texttt{LARGE}}$ Performance}
\label{sec:bartlarge}
\begin{figure}[!t]
\centering
\includegraphics[width=0.47\textwidth]{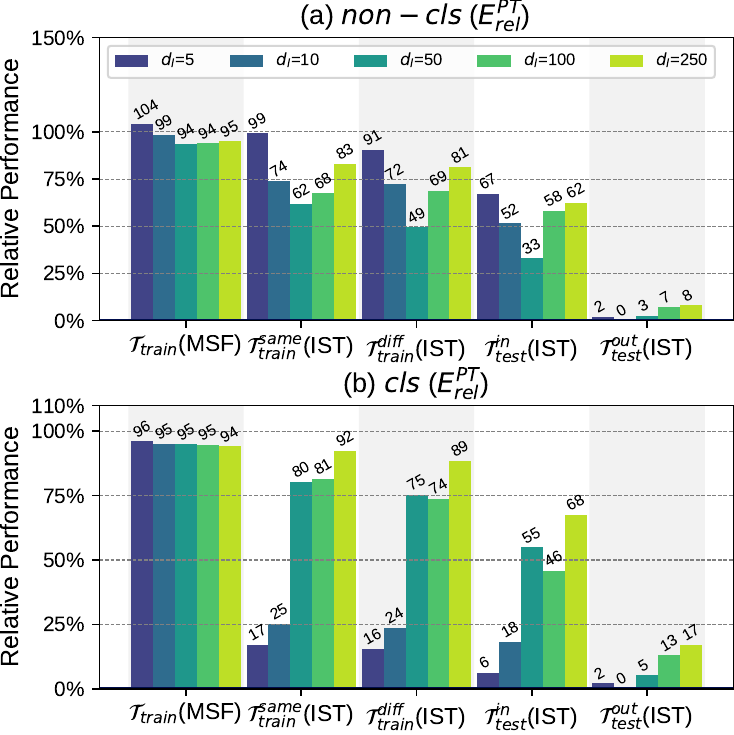}
\caption{Relative performance of \ourmodel with $\text{BART}_{\texttt{LARGE}}$ at different dimension $d_{I}$ on \textit{non-cls} and \textit{cls} splits, compared with prompt tuning.
}
\label{fig:large}
\end{figure}

All the experiments in \cref{sec:exp} are conducted with $\text{BART}_{\texttt{BASE}}$ model~\citep{lewis-etal-2020-bart}, which is also the main evaluated model of our adopted evaluation platform \textit{CrossFit}~\citep{ye2021crossfit}. %It is also interesting to see whether the conclusions will also hold for larger models.
To see whether the conclusions will also hold for larger models, we take a prior trial by conducting experiments on $\text{BART}_{\texttt{LARGE}}$. %and the results are shown in Figure~\ref{fig:large}. 
As the results in Figure~\ref{fig:large} suggest, the overall conclusions are consistent with those of $\text{BART}_{\texttt{BASE}}$ that non-trivial performance can be recovered in the found subspaces. However, when $d_I$ is extremely low ($5\sim10$), the performance is obviously worse than the cases of $\text{BART}_{\texttt{BASE}}$, especially on the \textit{cls} split. This phenomenon may come from the difficulty of finding intrinsic task subspaces for larger PLMs. Considering that \citet{aghajanyan-etal-2021-intrinsic} indicate that larger PLMs generally have smaller intrinsic dimensions, we encourage future work to investigate how to better find the intrinsic task subspace for PLMs with larger sizes at small intrinsic dimensions. Since our main focus is to demonstrate the existence of the universal intrinsic task subspace and advance further research explorations, in this paper, we do not experiment on other kinds of PLMs or PLMs with extremely large sizes.

\subsection{Comparison of Subspace-Finding Methods on $\mathcal{T}_{\mathrm{train}}$}
\label{sec:random_training}

\begin{table}[!t]
  \centering
  \small
    \begin{tabular}{ccccc}
    \toprule
     \multicolumn{1}{c}{\multirow{2}[2]{*}{\textbf{Split}}} & \multicolumn{2}{c}{\textbf{Prompt Tuning}} & \multicolumn{2}{c}{\textbf{Fine-tuning}} \\
 & \multicolumn{1}{c}{$\mathcal{T}_\text{train}$} & \multicolumn{1}{c}{$\mathcal{T}_\text{test}^{\text{in}}$ / $\mathcal{T}_\text{test}^{\text{out}}$} &
          \multicolumn{1}{c}{$\mathcal{T}_\text{train}$} & \multicolumn{1}{c}{$\mathcal{T}_\text{test}^{\text{in}}$ / $\mathcal{T}_\text{test}^{\text{out}}$} \\
    \midrule
\textit{random} &  $32.6$    &   $40.1$  ($\mathcal{T}_\text{test}$)    &  $35.2$     &  $40.7$ ($\mathcal{T}_\text{test}$) \\
\textit{non-cls} &  $23.0$    &   $28.0$ / $49.0$    &   $24.4$    &  $29.6$ / $52.2$ \\
\textit{cls} &  $48.6$    &  $50.9$ / $25.7$     & $52.5$      &  $51.1$ / $27.2$ \\
    \bottomrule
    \end{tabular}
  \caption{Average absolute performance for prompt tuning / fine-tuning on the three task splits we adopted.
  }
  \label{tab:abs_pt_ft}
\end{table}

\begin{table}[!t]
\small
  \centering
    \begin{tabular}{lccccc}
    \toprule
    \textbf{Dim} ($d_I$) & $5$     & $10$    & $50$    & $100$ & $250$     \\
    \midrule
    \multicolumn{6}{l}{\textit{Multi-task Subspace Finding}}\\
    $\mathcal{T}_\text{train}$ & $31.8$ & $32.2$  & $32.0$    & $32.6$  & $\mathbf{33.0}$ \\
    $\mathcal{T}_\text{test}$ & $10.0$  & $15.0$ & $\mathbf{16.7}$ & $16.4$ & $16.6$\\
    \midrule
    \multicolumn{6}{l}{\textit{Intrinsic Subspace Tuning}}\\
    $\mathcal{T}_\text{train}^\text{same}$  & $27.9$ & $32.1$ & $33.0$ & $\mathbf{33.0}$  & $31.7$\\
    $\mathcal{T}_\text{train}^\text{diff}$    & $28.1$ & $31.2$ & $31.2$ & $\mathbf{32.1}$  & $32.0$ \\
    $\mathcal{T}_\text{test}$ & $28.0$  & $26.8$ & $24.0$ & $26.3$ & $\mathbf{33.7}$ \\
    \bottomrule
    \end{tabular}
  \caption{Average absolute performance on the \textit{random} task split.}
  \label{tab:abs_random}
\end{table}

\begin{table}[!t]
\small
  \centering
    \begin{tabular}{lccccc}
    \toprule
    \textbf{Dim} ($d_I$)   & $5$     & $10$    & $50$    & $100$ & $250$\\
    \midrule
    \multicolumn{6}{l}{\textit{Multi-task Subspace Finding}}\\
    $\mathcal{T}_\text{train}$ & $23.1$ & $21.9$ & $22.7$ & $22.2$  & $\mathbf{22.7}$\\
    \midrule
    \multicolumn{6}{l}{\textit{Intrinsic Subspace Tuning}}\\
    $\mathcal{T}_\text{train}^\text{same}$  & $\mathbf{27.4}$ & $23.4$ & $27.2$ & $26.2$  & $26.8$\\
    $\mathcal{T}_\text{train}^\text{diff}$ & $25.8$ & $22.8$ & $25.4$ & $25.8$ & $\mathbf{26.6}$\\
    $\mathcal{T}_\text{test}^\text{in}$ & $17.4$ & $17.7$ & $18.6$ & $21.5$  & $\mathbf{24.2}$\\
    $\mathcal{T}_\text{test}^\text{out}$      & $1.0$  & $0.8$  & $1.4$  & $3.9$ & $\mathbf{15.8}$ \\
    \bottomrule
    \end{tabular}
  \caption{Average absolute performance on the \textit{non-cls} task split.}
  \label{tab:abs_non_cls}
\end{table}

\begin{table}[!t]
\small
  \centering
    \begin{tabular}{lccccc}
    \toprule
    \textbf{Dim} ($d_I$)   & $5$     & $10$    & $50$    & $100$  & $250$\\
    \midrule
    \multicolumn{6}{l}{\textit{Multi-task Subspace Finding}}\\
    $\mathcal{T}_\text{train}$ & $\mathbf{50.0}$    & $48.0$    & $49.5$  & $48.7$  & $46.0$\\
    \midrule
    \multicolumn{6}{l}{\textit{Intrinsic Subspace Tuning}}\\
    $\mathcal{T}_\text{train}^\text{same}$  & $35.2$    & $31.9$  & $\mathbf{51.0}$  & $49.1$ & $47.9$\\
    $\mathcal{T}_\text{train}^\text{diff}$  & $34.3$    & $31.9$  & $\mathbf{49.7}$  & $46.2$ & 48.6 \\
    $\mathcal{T}_\text{test}^\text{in}$ & $21.0$    & $24.5$  & $32.7$  & $\mathbf{38.1}$ & $35.0$ \\
    $\mathcal{T}_\text{test}^\text{out}$   & $0.7$   & $1.0$     & $2.3$   &  $4.6$ & $\mathbf{7.4}$ \\
    \bottomrule
    \end{tabular}
  \caption{Average absolute performance on the \textit{cls} task split.}
  \label{tab:abs_cls}
\end{table}

\begin{figure}[!th]
\centering
\includegraphics[width=0.47\textwidth]{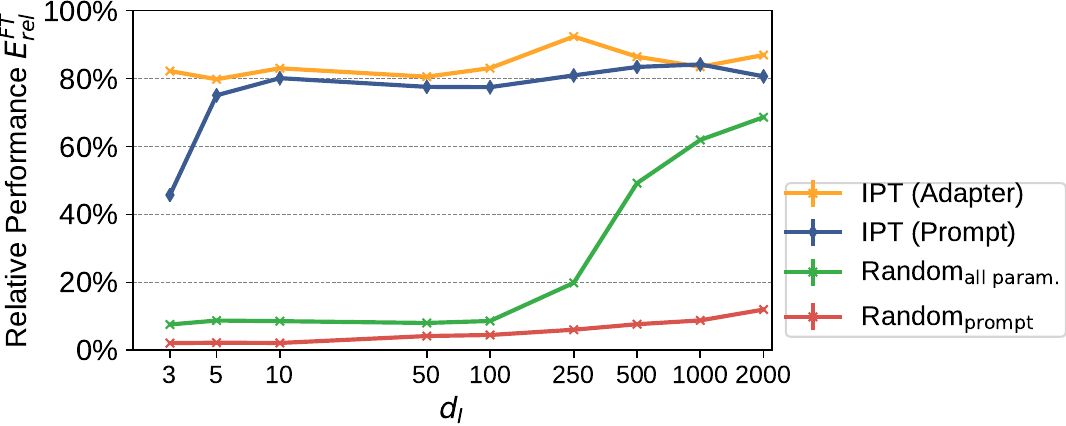}
\caption{Comparisons between \ourmodel and other subspace-finding methods on $\mathcal{T}_{\mathrm{train}}$ of \textit{random} task split.
}
\label{fig:vsrandom_train}
\end{figure}

In \cref{exp:analysis}, we compare \ourmodel and other subspace-finding methods on $\mathcal{T}_{\mathrm{test}}$ of \textit{random} split, i.e., the \textit{unseen task} challenge. Here we additionally compare them on the \textit{unseen data} challenge, i.e., $\mathcal{T}_{\mathrm{train}}$. The results are shown in Figure~\ref{fig:vsrandom_train}, from which we can see that: (1) the methods relying on a random subspace (Random$_{\mathrm{prompt}}$ and Random$_{\mathrm{all~param.}}$) do not involve training and thus their performance is poorer than our \ourmodel, especially at low dimensions; (2) for IPT(Prompt) and IPT(Adapter), their performance is consistently high at each dimension, which suggests that our pipeline works well in memorizing (fitting) the reparameterization of training tasks.

\section{Additional Visualization}
\label{sec:visual_2}
\begin{figure}[!t]
\centering
\includegraphics[width=0.45\textwidth]{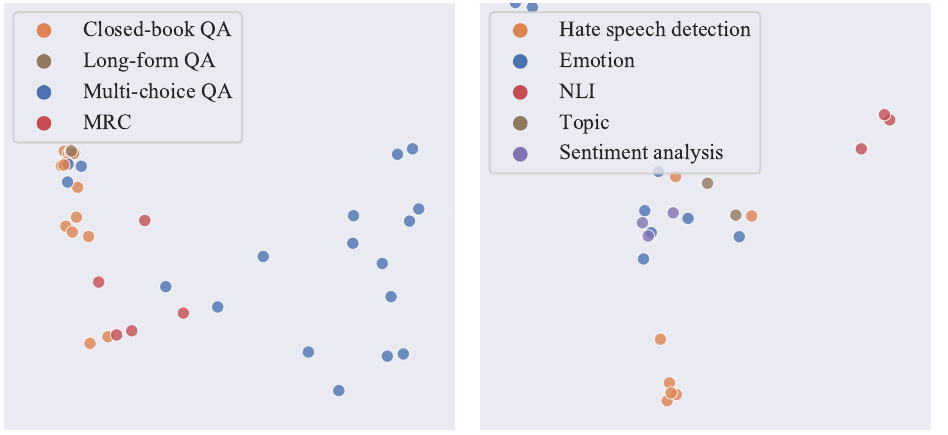}
\caption{PCA plots of the intrinsic vectors learned during IST. We label points with different colors to represent their corresponding categories. Specifically, we show the clusters of fine-grained categories of QA (left) and text classification tasks (right). Without loss of generality, we choose the task split of \textit{random} and $d_I = 100$.}
\label{fig:visualization_2}
\end{figure}

We visualize the intrinsic vectors of fine-grained categories of QA and text classification tasks using PCA in Figure \ref{fig:visualization_2}. We observe that points belonging to the same category exhibit a compact cluster. This further shows that the learned intrinsic vectors could serve as task representations and help us analyze the similarity and differences of diverse NLP tasks.

\section{Descriptions of Notations}
\label{sec:notation}
We explain some of the notations used in this paper in Table~\ref{tab:notation}, so that readers could more easily navigate through the whole paper.

\begin{table*}[!t]
\small
  \centering
    \begin{tabular}{cl}
    \toprule
    \textbf{Notation}   & \multicolumn{1}{c}{\textbf{Description}} \\
    \midrule
    $\text{IPT}$ & The analysis pipeline proposed in this paper (Intrinsic Prompt Tuning).  \\
    $\text{MSF}$ & The first stage of IPT (Multi-task Subspace Finding).  \\
    $\text{IST}$ & The second stage of IPT (Intrinsic Subspace Tuning).  \\
    \midrule
    $\mathcal{T}_\text{train}$(MSF) & Reconstructing the trained soft prompts of training tasks by training an auto-encoder. \\
    $\mathcal{T}_\text{test}$(MSF) & Testing the generalization of the trained auto-encoder on test tasks. \\
    $\mathcal{T}_\text{train}^{\text{same}}$(IST) & Conducting IST for each training task with the same training data used in MSF.  \\
    $\mathcal{T}_\text{train}^{\text{diff}}$(IST) & Conducting IST for each training task with the different training data used in MSF (\textit{unseen-data challenge}).  \\
    $\mathcal{T}_\text{test}$(IST) & Conducting IST for each test task (\textit{unseen-task challenge}). \\
    $\mathcal{T}_\text{test}^{\text{in}}$(IST) & Conducting IST for each test task belonging to the same categories of training tasks (\textit{unseen-task challenge}). \\
    $\mathcal{T}_\text{test}^{\text{out}}$(IST) & Conducting IST for each test task belonging to different categories of training tasks (\textit{unseen-task challenge}). \\
    \midrule
    $\mathcal{D}^i_\text{train}$ & The training set of the task $\mathcal{T}_i$. \\
    $\mathcal{D}^i_\text{dev}$ & The development set of the task $\mathcal{T}_i$. \\
    $\mathcal{D}^i_\text{test}$ & The test set of the task $\mathcal{T}_i$. \\
    \midrule
    $E_{\text{rel}}^\text{FT}$ & Average relative performance compared with fine-tuning (FT). \\
    $E_{\text{rel}}^\text{PT}$ & Average relative performance compared with prompt tuning (PT). \\
    \midrule
    \textit{random} & Randomly split tasks $\mathcal{T}_\text{all}$ into training tasks $\mathcal{T}_\text{train}$ and test tasks $\mathcal{T}_\text{test}$. \\
    \textit{non-cls} & Choose a subset of non-classification tasks as the training tasks $\mathcal{T}_\text{train}$. \\
    \textit{cls} & Choose a subset of classification tasks as the training tasks $\mathcal{T}_\text{train}$. \\
    \bottomrule
    \end{tabular}
  \caption{Descriptions about the notations used in this paper.}
  \label{tab:notation}
\end{table*}

\section{Implementation Details}
\label{sec:training_detail}

As mentioned in \cref{sec:preliminary}, all tasks are processed into a unified sequence-to-sequence format following \citet{raffel2020exploring} and \citet{khashabi-etal-2020-unifiedqa} for ease of handling them with unified text-to-text PLMs.

\subsection{\ourmodel Details}
For all experiments, we adopt AdamW \citep{loshchilov2017decoupled} as the optimizer. We train all models under the same environment of NVIDIA 32GB V100 GPU. We perform grid search on the combination of a series of learning rates ($\{1\times10^{-5}, 2\times10^{-5}, 5\times10^{-5}, 1\times10^{-4}\}$) and batch sizes ($\{2,4,8\}$)\footnote{The numbers are chosen by pilot experiments on a random subset of tasks.}, choose the best checkpoint using $\mathcal{D}_\text{dev}$, and evaluate it on $\mathcal{D}_\text{test}$. We set the max step to $10,000$ / $100,000$ and validate on $\mathcal{D}_\text{dev}$ every $100$ / $1000$ steps\footnote{We found that prompt tuning empirically requires around $10\times$ more training steps than fine-tuning to converge.}. The ratio $\alpha$ is set to $200$. The hyper-parameters of IST are chosen as the same as prompt tuning for fair comparisons. To avoid possible information leakage, we discard the low-dimensional solutions (intrinsic vectors) found during MSF, and randomly initialize the intrinsic vector before conducting IST. The experiments of MSF could be finished within $24$ hours using $8$ GPUs, and each experiment in IST could be finished within $6$ hours on average using $1$ GPU.
% During MSF, we only select the prompts that perform best on $\mathcal{D}_\text{dev}$ for each task to train the auto-encoder since we empirically found that involving other prompts leads to worse performance.

For detailed model implementation, as mentioned in \cref{sec:IPT}, the projection function $\mathbf{Proj}(\cdot)$ is implemented with a one-layer feed-forward network and $\mathbf{Proj}_{b}(\cdot)$ is parameterized by a two-layer perceptron as follows:
\begin{equation}
\begin{aligned}
\mathbf{Proj}_{b}(\mathbf{d}_i) = \mathbf{W}_2(\tanh(\mathbf{W}_1\mathbf{d}_i + \mathbf{b}_1)) + \mathbf{b}_2, \nonumber
\end{aligned}
\end{equation}
where $\mathbf{W}_1 \in \mathbb{R}^{d_I' \times d_I}$, $\mathbf{b}_1 \in \mathbb{R}^{d_I'}$, $\mathbf{W}_2 \in \mathbb{R}^{n \times d \times d_I'}$ and $\mathbf{b}_2 \in \mathbb{R}^{n \times d}$ are trainable parameters. $d_I$ denotes the intrinsic dimension investigated in this paper. We set the inner hidden size $d_I'$ of $\mathbf{Proj}_{b}$ to $768$ for both $\text{BART}_{\texttt{BASE}}$ and $\text{BART}_{\texttt{LARGE}}$.

\subsection{Subspace-Finding with Adapter}
\label{sec:training_detail_adapter}
The proposed \ourmodel is agnostic to the specific parameter-efficient tuning method. To demonstrate this, we apply the techniques of \ourmodel to Adapter~\citep{Houlsby2019Adapter}, which is a representative parameter-efficient tuning algorithm.

\begin{figure*}[!t]
    \centering
	\includegraphics[width=0.85\textwidth]{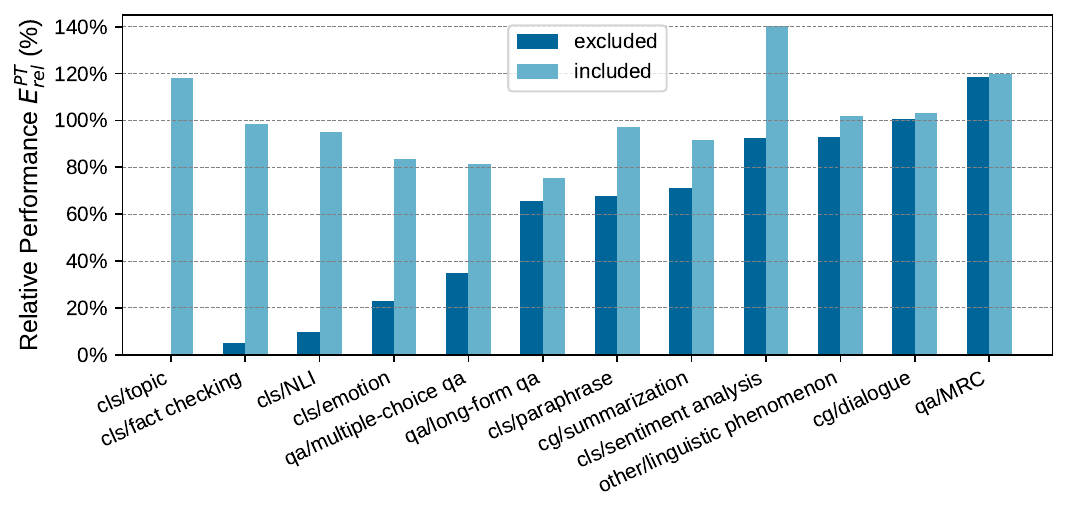}
    \caption{$E^{\text{PT}}_{\text{rel}}$ performance of \ourmodel on tasks grouped by fine-grained task types. \textit{included} means that MSF is conducted on the training tasks of the \textit{random} split (including all the task types). \textit{excluded} means that MSF is conducted on the training tasks of the \textit{random} split excluding the tasks of the investigated type. Note the performance of IST ($E^{\text{PT}}_{\text{rel}}$) is tested on the tasks of the investigated type.}
    \label{fig:task_ontology_ipt}
\end{figure*}

Adapter plugs in lightweight feed-forward networks between Transformer layers (both after the MHA module and the FFN module). Every adapter module consists of a down-projection matrix $\mathbf{W}_{\text{down}}\in\mathbb{R}^{r_\text{A}\times d}$, 
a non-linear activation function $f(\cdot)$, and a up-projection matrix 
$\mathbf{W}_{\text{up}}\in\mathbb{R}^{d\times r_\text{A}}$, 
where $r_\text{A}$ denotes the bottleneck dimension, and $d$ denotes the hidden size of the PLM. Given an input $\mathbf{h}$, adapter applies a residual connection as follows:
\begin{align*}
% \label{eq:adapter}
    \mathbf{h}\leftarrow \mathbf{h} + \mathbf{W}_{\text{up}}f(\mathbf{W}_{\text{down}}\mathbf{h}).
\end{align*}

The essence of \ourmodel is to define a projector from the intrinsic task subspace to the original parameter space. For Adapter, the parameter space is decided by the newly introduced matrices $\mathbf{W}_{\text{down}}$ and $\mathbf{W}_{\text{up}}$ in each layer. During the first stage MSF, we reparameterize both $\mathbf{W}_{\text{down}}$ and $\mathbf{W}_{\text{up}}$ as the product of a low-dimensional intrinsic vector $\mathbf{V} \in \mathbb{R}^{d_I}$, lying in the intrinsic task subspace, and the corresponding projection matrices as follows:
\begin{align*}
    (\mathbf{W}_{\text{up}}; \mathbf{W}_{\text{down}}) &= \textbf{Proj} (\mathbf{V}),
\end{align*}
where the projection \textbf{Proj} is implemented by a two-layer perceptron. During the second stage of IST, only a $d_I$-dimensional (randomly initialized) intrinsic vector $\mathbf{V}$ is optimized. Other implementation details are kept the same as IPT(Prompt).

\section{Fine-grained Analyses on Task Types}
\label{sec:task_type}

In \cref{sec:exp}, we evaluate the performance of \ourmodel on $120$ tasks and also divide them into cls. (classification) and non-cls. (non-classification) tasks to see the difference between these two types. Here we take a step further to investigate \text{IPT} at a more fine-grained level based on the task ontology of~\citet{ye2021crossfit}. Specifically, we choose $6$ cls. task types (cls/topic, cls/fact checking, cls/NLI, cls/emotion, cls/paraphrase, cls/sentiment analysis) and $6$ non-cls. task types (qa/multiple-choice qa, qa/long-form qa, cg/summarization, other/linguistic phenomenon, cg/dialogue, qa/MRC). We report the relative performance of \ourmodel compared with prompt tuning ($E^{\text{PT}}_{\text{rel}}$) on these fine-grained task types, including two settings:

$\bullet$ \textit{included} means that MSF is conducted on the training tasks of the \textit{random} split ($100$ tasks in total, including all the task types), and IST is conducted on the tasks of the investigated type;

$\bullet$ \textit{excluded} means that MSF is conducted on the training tasks excluding the tasks of the investigated type. In other words, we exclude the investigated type of tasks from the training tasks, to form the new training task set. The experiments of IST are conducted using the tasks of the investigated type. The test tasks of both settings (\textit{included} and \textit{excluded}) are the same.

Intuitively, if we exclude a specific task type from the training tasks, the approximated intrinsic task subspace may fail to include the language skills required by this task type, and would thus result in poor recovering performance during IST on the investigated task type. However, if other tasks in the training set share similar language skills of the investigated task type, the above issue could be mitigated to some extent. In this sense, the performance gap between \textit{excluded} and \textit{included} of a task type reflects the relation (common language skills) of the investigated type to other task types.

From the results shown in Figure~\ref{fig:task_ontology_ipt}, we can observe that: (1) some task types have a significant distinction from other types (reflected in the huge gap between \textit{excluded} and \textit{included} performance), such as cls/topic, cls/fact checking, and cls/NLI, which may come from the unique skills required to solve these tasks; (2) instead, some tasks tend to require similar language skills than other task types (reflected in the close performance between \textit{excluded} and \textit{included}), such as other/linguistic phenomenon, cg/dialogue, and qa/MRC. The above results demonstrate the potential of \ourmodel to analyze task differences and we encourage future works to explore it systematically; (3) \ourmodel achieves obvious improvements compared to vanilla prompt tuning (i.e., $E^{\text{PT}}_{\text{rel}} > 100\%$) on some task types such as cls/topic and qa/MRC, which indicates that tuning PLMs within the intrinsic task subspace is promising to obtain performance benefits.

Taking a step further, the above results also raise interesting research questions: (1) first of all, can we break the task barrier and decompose each task into a series of ``atom tasks''? (2) Does there exist a limited number of atom task set, building on which all NLP tasks can be composed? Intuitively, solving each atom task requires a specific skill of neural models. By splitting each task into corresponding atom tasks and correctly defining the required skills of each atom task, we can better understand the characteristics of each task and fathom the relation among different tasks. Our analyses in this paper could provide supporting facts for the above questions. For instance, one can view each dimension of our intrinsic subspace as one kind of skill PLMs own, and solving each task can be seen as stimulating and composing the latent skills stored in PLMs. We encourage future works to investigate these interesting research questions.

\section{Task Details}
We list details for all the evaluated tasks in this paper in Table~\ref{tab:ontology_1}.
\label{sec:ontology}
\begin{table*}[!h]
  \centering
  \small
  \caption{The tasks evaluated in our experiments. We refer to \citet{ye2021crossfit} for task ontology.}
    \begin{tabular}{clll}
    \toprule
    \textbf{Ontology} & \textbf{Task Name} & \textbf{Reference}\\
    \midrule
    \multirow{3}[2]{*}{cls/sentiment analysis} 
          & glue-sst2 & \citealt{socher-etal-2013-recursive} \\
          & imdb  & \citealt{maas-etal-2011-learning}   \\
          & rotten\_tomatoes & \citealt{pang-lee-2005-seeing} \\
    \midrule
    \multirow{10}[2]{*}{cls/emotion} 
          & emo & \citealt{chatterjee-etal-2019-semeval}    \\
          & tweet\_eval-emoji & \citealt{barbieri-etal-2020-tweeteval}    \\
          & tweet\_eval-hate &	\citealt{barbieri-etal-2020-tweeteval} \\
          & tweet\_eval-irony &	\citealt{barbieri-etal-2020-tweeteval} \\
          & tweet\_eval-offensive &	\citealt{barbieri-etal-2020-tweeteval} \\
          & tweet\_eval-sentiment  &	\citealt{barbieri-etal-2020-tweeteval} \\
          & tweet\_eval-stance\_abortion &	\citealt{barbieri-etal-2020-tweeteval} \\
          & tweet\_eval-stance\_atheism &	\citealt{barbieri-etal-2020-tweeteval} \\
          & tweet\_eval-stance\_climate &	\citealt{barbieri-etal-2020-tweeteval} \\
          & tweet\_eval-stance\_feminist &	\citealt{barbieri-etal-2020-tweeteval} \\
          & tweet\_eval-stance\_hillary &	\citealt{barbieri-etal-2020-tweeteval} \\
    \midrule
    \multirow{7}[2]{*}{cls/hate speech detection} 
          &   ethos-disability   &	\citealt{Mollas2020ETHOSAO} \\
          &   ethos-gender   &	\citealt{Mollas2020ETHOSAO} \\
          &   ethos-national\_origin &	\citealt{Mollas2020ETHOSAO}   \\
          &   ethos-religion   &	\citealt{Mollas2020ETHOSAO} \\
          &   ethos-sexual\_orientation   &	\citealt{Mollas2020ETHOSAO} \\
          &   hate\_speech18  &	\citealt{hateoffensive} \\
          &   hatexplain &	\citealt{mathew2020hatexplain}  \\
    \midrule
    \multirow{7}[2]{*}{cls/NLI} 
          &   anli & \citealt{nie-etal-2020-adversarial}  \\
          &  glue-mnli & \citealt{williams-etal-2018-broad}   \\
          &  glue-qnli &	\citealt{rajpurkar-etal-2016-squad}   \\
          &  glue-rte  &	\begin{tabular}[c]{@{}l@{}}\citealt{dagan2005pascal, bar2006second}\\\citealt{giampiccolo2007third, bentivogli2009fifth}\end{tabular}  \\
          &  glue-wnli &	\citealt{faruqui-das-2018-identifying}   \\
          &  scitail  & \citealt{Khot2018SciTaiLAT}  \\
          &  superglue-rte & \begin{tabular}[c]{@{}l@{}}\citealt{dagan2005pascal, bar2006second}\\\citealt{giampiccolo2007third, bentivogli2009fifth}\end{tabular}   \\
    \midrule
    \multirow{3}[2]{*}{cls/fact checking} 
          &  climate\_fever  &	\citealt{Diggelmann2020CLIMATEFEVERAD}  \\
          &   kilt\_fever &	\citealt{thorne-etal-2018-fever}  \\
          &   liar  &	\citealt{wang-2017-liar} \\
    \midrule
    \multirow{3}[2]{*}{cls/paraphrase} 
          &   glue-qqp & \href{https://quoradata.quora.com/First-Quora-Dataset-Release-Question-Pairs}{(link)}  \\
          &  medical\_questions\_pairs & \citealt{medical-qqp}  \\
          &  paws & \citealt{zhang-etal-2019-paws}   \\
    \midrule
    \multirow{2}[2]{*}{cls/topic} 
          &   ag\_news  &	\href{http://groups.di.unipi.it/~gulli/AG_corpus_of_news_articles.html}{Gulli (link)} \\
          &   dbpedia\_14 &	\citealt{Lehmann2015DBpediaA}  \\
    \midrule
    \multirow{8}[2]{*}{cls/other} 
          &  ade\_corpus\_v2-classification &	\citealt{GURULINGAPPA2012885}   \\
          &   discovery  &	\citealt{sileo-etal-2019-mining} \\
          &   glue-cola  &	\citealt{warstadt-etal-2019-neural} \\
          &   google\_wellformed\_query &	\citealt{faruqui-das-2018-identifying}  \\
          &   sms\_spam &	\citealt{sms_spam}   \\
          &   superglue-wic  &	\citealt{pilehvar-camacho-collados-2019-wic} \\
          &   superglue-wsc &	\citealt{levesque2012winograd}  \\
          &   wiki\_qa &	\citealt{yang-etal-2015-wikiqa}  \\
    \midrule
    \multirow{10}[2]{*}{qa/closed-book qa} 
          & freebase\_qa  &	\citealt{jiang-etal-2019-freebaseqa}    \\
          &  jeopardy &	\href{https://www.reddit.com/r/datasets/comments/1uyd0t/200000_jeopardy_questions_in_a_json_file/}{(link)}   \\
          &  kilt\_hotpotqa  &	\citealt{yang-etal-2018-hotpotqa}  \\
          &  kilt\_nq  &	\citealt{kwiatkowski-etal-2019-natural}  \\
          &  kilt\_trex  &	\citealt{elsahar-etal-2018-rex}  \\
          &  kilt\_zsre &	\citealt{levy-etal-2017-zero}   \\
          &  lama-conceptnet &	\citealt{petroni-etal-2019-language,petroni2020how}   \\
          &  lama-google\_re &	\citealt{petroni-etal-2019-language,petroni2020how}   \\
          &  lama-squad &	\citealt{petroni-etal-2019-language,petroni2020how}    \\
          &  lama-trex &	\citealt{petroni-etal-2019-language,petroni2020how}   \\
          &  numer\_sense &	\citealt{lin-etal-2020-birds}   \\
          &  search\_qa  &	\citealt{Dunn2017SearchQAAN}  \\
          &  squad-no\_context  &	\citealt{rajpurkar-etal-2016-squad}  \\
          &  web\_questions  &	\citealt{berant-etal-2013-semantic}  \\
    \midrule
    \multirow{2}[2]{*}{qa/binary} 
          &  boolq &	\citealt{clark-etal-2019-boolq}   \\
          &  mc\_taco  &	\citealt{zhou-etal-2019-going}  \\
    \bottomrule
    \end{tabular}%
  \label{tab:ontology_1}%
\vspace{-2em}
\end{table*}%

\begin{table*}[b]
  \centering
  \small
    \begin{tabular}{clll}
    \toprule
    \textbf{Ontology} & \textbf{Task Name} 
          & \textbf{Reference}\\
    \midrule
    \multirow{21}[2]{*}{qa/multiple-choice qa} 
          &  ai2\_arc &	\citealt{Clark2018ThinkYH}   \\
          &  aqua\_rat &	\citealt{ling-etal-2017-program}  \\
          &  codah  &	\citealt{chen-etal-2019-codah}  \\
          &  commonsense\_qa &	\citealt{talmor-etal-2019-commonsenseqa}   \\
          &  cosmos\_qa  &	\citealt{huang-etal-2019-cosmos}  \\
          &  dream &	\citealt{saha-etal-2018-duorc}   \\
          &  hellaswag &	\citealt{zellers-etal-2019-hellaswag}   \\
          &  math\_qa &	\citealt{amini-etal-2019-mathqa}   \\
          &  openbookqa &	\citealt{mihaylov-etal-2018-suit}   \\
          &  qasc  &	\citealt{khot2020qasc}  \\
          &  quail  &	\citealt{Rogers_Kovaleva_Downey_Rumshisky_2020}  \\
          &  quarel &	\citealt{Tafjord_Clark_Gardner_Yih_Sabharwal_2019}   \\
          &  quartz-no\_knowledge &	\citealt{tafjord-etal-2019-quartz}   \\
          &  quartz-with\_knowledge &	\citealt{tafjord-etal-2019-quartz}   \\
          &  race-high &	\citealt{lai-etal-2017-race}   \\
          &  race-middle &	\citealt{lai-etal-2017-race}   \\
          &  social\_i\_qa  &	\citealt{sap-etal-2019-social}  \\
          &  superglue-copa  &	\citealt{gordon-etal-2012-semeval}  \\
          &  superglue-multirc &	\citealt{khashabi-etal-2018-looking}   \\
          &  swag  &	\citealt{zellers-etal-2018-swag}  \\
          &  wino\_grande &	\citealt{Sakaguchi_Le_Bras_Bhagavatula_Choi_2020}   \\
    \midrule
    \multirow{3}[2]{*}{qa/long-form qa} 
          & eli5-askh   & \citealt{fan-etal-2019-eli5}   \\
          &  eli5-asks & \citealt{fan-etal-2019-eli5}   \\
          &  eli5-eli5  & \citealt{fan-etal-2019-eli5}  \\
    \midrule
    \multirow{5}[2]{*}{qa/MRC} 
          &  adversarialqa &	\citealt{bartolo-etal-2020-beat}   \\
          &  biomrc  &	\citealt{pappas-etal-2020-biomrc}  \\
          &  quoref &	\citealt{dasigi-etal-2019-quoref}   \\
          &  ropes &	\citealt{lin-etal-2019-reasoning}   \\
          &  superglue-record  & \citealt{Zhang2018ReCoRDBT}  \\
    \midrule
    \multirow{4}[2]{*}{cg/summarization} 
          & gigaword  &	\citealt{napoles-etal-2012-annotated}     \\
          &  multi\_news & \citealt{fabbri-etal-2019-multi}   \\
          &  samsum &	\citealt{gliwa-etal-2019-samsum}   \\
          &  xsum &	\citealt{narayan-etal-2018-dont}   \\
    \midrule
    \multirow{2}[2]{*}{cg/dialogue} 
          &   empathetic\_dialogues & \citealt{rashkin-etal-2019-towards}  \\
          &  kilt\_wow &	\citealt{dinan2018wizard}   \\
    \midrule
    \multirow{4}[2]{*}{cg/other} 
          &  spider &	\citealt{yu-etal-2018-spider}   \\
          &  wiki\_bio &	\citealt{lebret-etal-2016-neural}   \\
          &  wiki\_split & \citealt{botha-etal-2018-learning}    \\
          &  wikisql &	\citealt{zhongSeq2SQL2017}   \\
    \midrule
    \parbox[c]{2cm}{other/linguistic phenomenon} 
          %&  \parbox[c]{4cm}{
          & blimp-anaphor\_gender\_agreement  & \citealt{warstadt2019blimp}  \\
          & blimp-ellipsis\_n\_bar\_1 & \citealt{warstadt2019blimp}   \\
          & blimp-sentential\_negation\_npi\_scope & \citealt{warstadt2019blimp} \\%}  \\
    % \multirow{3}[2]{*}{other/linguistic phenomenon} &  blimp-anaphor\_gender\_agreement    \\
    %       &  blimp-ellipsis\_n\_bar\_1    \\
    %       &  blimp-sentential\_negation\_npi\_scope    \\
    \midrule
    \parbox[c]{2cm}{other/generate explanation} 
          %&  \parbox[c]{4cm}{cos\_e}  \\
          & cos\_e & \citealt{rajani-etal-2019-explain}  \\
    \midrule
    \multirow{2}[2]{*}{other/slot\_filling} 
          &  ade\_corpus\_v2-dosage & \citealt{GURULINGAPPA2012885}   \\
          &  ade\_corpus\_v2-effect & \citealt{GURULINGAPPA2012885}   \\
    \midrule
    other/entity linking 
          &  kilt\_ay2 &	\citealt{hoffart-etal-2011-robust}   \\
    \midrule
    \multirow{14}[2]{*}{other/other} 
          &   acronym\_identification  & \citealt{pouran-ben-veyseh-etal-2020-acronym}  \\
          &  art &	\citealt{bhagavatula2020abductive}    \\
          &  aslg\_pc12  &	\citealt{Othman2012EnglishASLGP}  \\
          &  break-QDMR &	\citealt{wolfson-etal-2020-break}   \\
          &  break-QDMR-high-level  & \citealt{wolfson-etal-2020-break}  \\
          &  common\_gen  &	\citealt{lin-etal-2020-commongen}  \\
          &  crawl\_domain &	\citealt{zhang-etal-2020-semi}   \\
          &  crows\_pairs  &	\citealt{nangia-etal-2020-crows}  \\
          &  definite\_pronoun\_resolution &	\citealt{rahman-ng-2012-resolving}   \\
          &  e2e\_nlg\_cleaned &	\citealt{dusek.etal2020:csl, dusek-etal-2019-semantic}   \\
          &  limit  &	\citealt{manotas-etal-2020-limit}  \\
          &  piqa  &	\citealt{Bisk2020}   \\
          &  proto\_qa &	\citealt{boratko-etal-2020-protoqa}   \\
          &  qa\_srl &	\citealt{he-etal-2015-question}   \\
    \bottomrule
    \end{tabular}%
  \label{tab:ontology_2}%
\end{table*}%

\end{document}